\documentclass[final]{cvpr}

\usepackage{times}
\usepackage{epsfig}
\usepackage{graphicx}
\usepackage{amsmath}
\usepackage{amssymb}
\usepackage{enumitem}
\usepackage{soul}
\usepackage[font=small,labelfont=small,bf]{caption}
\usepackage{url}
\usepackage[dvipsnames]{xcolor}
\usepackage{multicol}
\usepackage{multirow}
\setstcolor{red}
\usepackage{subcaption}

\newenvironment{tight_itemize}{
\begin{itemize}[leftmargin=15pt]
  \setlength{\topsep}{0pt}
  \setlength{\itemsep}{0pt}
  \setlength{\parskip}{0pt}
  \setlength{\parsep}{0pt}
}{\end{itemize}}

\usepackage{pifont}
\newcommand{\cmark}{\ding{51}}%
\newcommand{\xmark}{\ding{55}}%

\usepackage[pagebackref=true,breaklinks=true,colorlinks,bookmarks=false]{hyperref}

\def\cvprPaperID{10082}
\def\confYear{CVPR 2021}

\begin{document}
\title{OpenRooms: An Open Framework for Photorealistic Indoor Scene Datasets}

\author{ {Zhengqin Li$^{1}$}\quad 
{Ting-Wei Yu$^{1}$} \quad 
{Shen Sang$^{1}$}\quad 
{Sarah Wang$^{1}$}\quad 
{Meng Song$^{1}$}\quad 
{Yuhan Liu$^{1}$}\quad
{Yu-Ying Yeh$^{1}$}\quad \\
{Rui Zhu$^{1}$}\quad
{Nitesh Gundavarapu$^{1}$}\quad
{Jia Shi$^{1}$}\quad
{Sai Bi$^{1}$}\quad
{Hong-Xing Yu$^{1}$}\quad
{Zexiang Xu$^{2}$}\quad \\
{Kalyan Sunkavalli$^{2}$}\quad
{Milo\v{s} Ha\v{s}an$^{2}$}\quad 
{Ravi Ramamoorthi$^{1}$}\quad
{Manmohan Chandraker$^{1}$} \\[2mm]
{$^{1}$UC San Diego }\quad {$^{2}$Adobe Research}
}

\maketitle

\begin{abstract}
We propose a novel framework for creating large-scale photorealistic datasets of indoor scenes, with ground truth geometry, material, lighting and semantics. Our goal is to make the dataset creation process widely accessible, transforming scans into photorealistic datasets with high-quality ground truth for appearance, layout, semantic labels, high quality spatially-varying BRDF and complex lighting, including direct, indirect and visibility components. This enables important applications in inverse rendering, scene understanding and robotics. We show that deep networks trained on the proposed dataset achieve competitive performance for shape, material and lighting estimation on real images, enabling photorealistic augmented reality applications, such as object insertion and material editing. We also show our semantic labels may be used for segmentation and multi-task learning. Finally, we demonstrate that our framework may also be integrated with physics engines, to create virtual robotics environments with unique ground truth such as friction coefficients and correspondence to real scenes. The dataset and all the tools to create such datasets will be made publicly available.\footnote{Webpage: \url{https://ucsd-openrooms.github.io/}}
\end{abstract}

\vspace{-0.2cm}
\section{Introduction}
\label{sec:introduction}
\vspace{-0.1cm}

Indoor scenes represent important environments for visual perception and scene understanding, for applications such as augmented reality and robotics. However, their appearance is a complex function of multiple factors such as shape, material and lighting, and demonstrates phenomena like significant occlusions, shadows, interreflections and large spatial variations in lighting.
Reasoning about these underlying, entangled factors requires large-scale high-quality ground truth, which remains hard to acquire. While ground truth geometry can be captured using a 3D scanner, it is extremely challenging (if not nearly impossible) to accurately acquire the complex spatially-varying material and lighting of indoor scenes. An alternative is to consider synthetic datasets, but large-scale synthetic datasets of indoor scenes with plausible geometry, materials and lighting are also non-trivial to create.

This paper presents OpenRooms, a framework for synthesizing photorealistic indoor scenes, with broad applicability across computer vision, graphics and robotics. It has several advantages over prior works, summarized in Table \ref{tab:datasetComparison}. First, rather than use artist-created scenes and assets, we ascribe high-quality material and lighting to RGBD scans of real indoor scenes. Beyond just the data, we provide all the tools necessary to accomplish this, allowing any researcher to inexpensively create such datasets. While prior works can align CAD models to scanned point clouds \cite{avetisyan2019scan2cad, izadinia2017im2cad, avetisyan2020scenecad}, they do not explore how to assign materials and lighting appropriately to build a large-scale photorealistic dataset. Second, we provide extensive high-quality ground truth for complex light transport that is unmatched in prior works. Our material is represented by a spatially-varying microfacet bidirectional reflectance distribution function (SVBRDF), and our lighting includes windows, environment maps and area lights, along with their per-pixel spatially-varying effects to account for visibility, shadows and inter-reflections. Third, we render photorealistic images with our data and tools, which include a custom GPU-accelerated physically-based renderer.

\begin{figure*}[!!t]
\centering
\includegraphics[width=\textwidth]{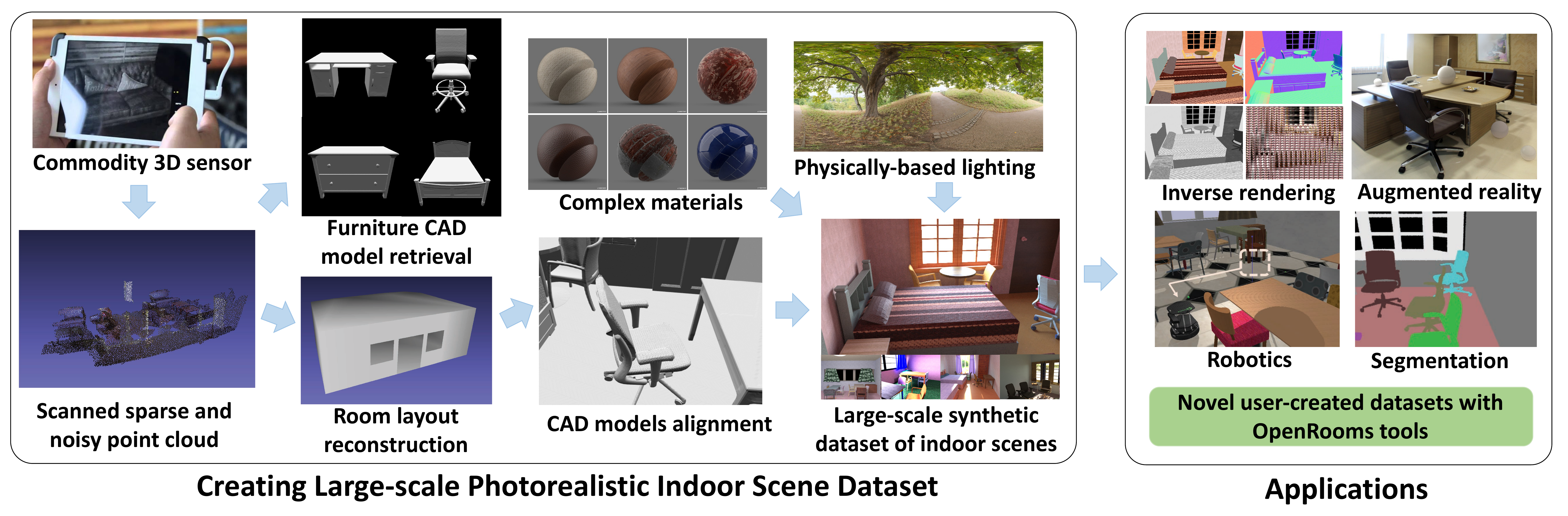}
\vspace{-0.4cm}
\caption{Our framework for creating a synthetic dataset of complex indoor scenes with ground truth shape, SVBRDF and SV-lighting, along with the resulting applications. Given possibly noisy scans acquired with a commodity 3D sensor, we generate consistent layouts for room and furniture. We ascribe per-pixel ground truth for material in the form of high-quality SVBRDF and for lighting as  spatially-varying physically-based representations. We render a large-scale dataset of images associated with this ground truth, which can be used to train deep networks for inverse rendering and semantic segmentation. We further motivate applications for augmented reality and robotics, while suggesting that the open source tools we make available can be used by the community to create other large-scale datasets too.}
\label{fig:teaser}
\vspace{-0.3cm}
\end{figure*}

We create an instance of such a dataset by building on existing repositories: 3D scans from ScanNet \cite{dai2017scannet}, CAD model alignment \cite{avetisyan2019scan2cad}, reflectance \cite{adobestock} and illumination \cite{HDRI,holdgeoffroy19sky}. The resulting dataset contains over 100K HDR images with ground-truth depths, normals, spatially-varying BRDF and light sources, along with per-pixel spatially-varying lighting and visibility masks for every light source. We also provide per-pixel semantic labels. Besides being publicly available, the dataset can be significantly extended through future community efforts based on our tools. We also demonstrate applicability of our method to other choices for material \cite{substance} and geometry \cite{song2015rgbd}.

We believe that our effort will significantly accelerate research in multiple areas. Inverse rendering tasks are directly related, including single-view \cite{eigen2015depth} and multi-view \cite{yao2018mvsnet} depth prediction, intrinsic decomposition \cite{li2018cgintrinsics,bi2018deep}, material classification \cite{bell2015minc} and lighting estimation \cite{gardner2017indoor,garon2019fast,li2020inverse}. To demonstrate the efficacy of the dataset, we train a state-of-the-art inverse rendering network and achieve accurate results on real images. We also demonstrate that OpenRooms may be used for training semantic segmentation networks \cite{zhao2017pyramid,chen2017rethinking}, as well as multi-task learning to jointly estimate shape, material and semantics. Our high-quality and extensive ground truth may help better understand complex light transport in indoor scenes and enable new applications in photorealistic augmented reality, where we demonstrate object insertion, material editing light source detection as examples, and may include light editing in the future.  

Studies in robotics may also benefit by using our ground truth to enhance existing simulation environments \cite{wu2018house3D,savva2017minos,xia2018gibson,habitat19iccv}.
We demonstrate this possibility by combining OpenRooms assets with the PyBullet engine \cite{PyBullet} and mapping our SVBRDFs to friction coefficients, to motivate navigation and rearrangement under different material and lighting. We also note that OpenRooms allows a one-to-one correspondence between real videos and simulations, which can be valuable for sim-to-real transfer  \cite{habitatsim2real20ral}.

In Figure~\ref{fig:teaser} we illustrate the OpenRooms framework for creating large-scale, high-quality synthetic indoor datasets from commodity RGBD sensor scans and demonstrate some of the applications that our work enables.

\begin{table*}[t]
\begin{minipage}[c]{0.75\textwidth}
\centering
\scriptsize
\setlength{\tabcolsep}{1pt}
\begin{tabular}{|c|cccccc|cccc|c|}
\hline
\multirow{2}{*}{Dataset} & \multicolumn{6}{|c|}{Available annotations} & \multicolumn{4}{|c|}{Publicly available assets} & \multirow{2}{*}{\shortstack[1]{Corresponding \\ real images \\ and scenes}} \\
\cline{2-11}
& Geometry  & Material  & \multicolumn{3}{c}{Lighting}              & Segmentation & Images & CAD & Baseline & Tool &  \\
\cline{4-6}
&           &           & Light sources & Per-pixel & Visibility    &              &        &     &          &      &  \\
\hline
PBRS \cite{zhang2016physically} & \color{OliveGreen}{\cmark} & \color{BurntOrange}{diffuse} & \color{red}{\xmark} & \color{BurntOrange}{shading} & \color{red}{\xmark} & \color{OliveGreen}{\cmark} & \color{red}{\xmark} & \color{red}{\xmark} & \color{OliveGreen}{\cmark} &  \color{OliveGreen}{\cmark} & \color{red}{\xmark} \\ 
\hline 
Scenenet \cite{mccormac2017scenenetrgbd}  & \color{OliveGreen}{\cmark} & \color{red}{\xmark} & \color{red}{\xmark} & \color{red}{\xmark} & \color{red}{\xmark} & \color{OliveGreen}{\cmark} & \color{OliveGreen}{\cmark} & \color{OliveGreen}{\cmark} & \color{red}{\xmark}  &  \color{OliveGreen}{\cmark} & \color{red}{\xmark} \\
\hline
CGIntrinsic \cite{li2018cgintrinsics} & \color{red}{\xmark} & \color{BurntOrange}{diffuse} & \color{red}{\xmark} & \color{BurntOrange}{shading} & \color{red}{\xmark} & \color{red}{\xmark} & \color{OliveGreen}{\cmark} & \color{red}{\xmark} & \color{OliveGreen}{\cmark} &  \color{OliveGreen}{\cmark} & \color{red}{\xmark} \\ 
\hline
InteriorNet \cite{li2018interiornet} & \color{OliveGreen}{\cmark} & \color{BurntOrange}{diffuse} & \color{red}{\xmark} & \color{BurntOrange}{shading} & \color{red}{\xmark} & \color{OliveGreen}{\cmark} & \color{OliveGreen}{\cmark} & \color{red}{\xmark} & 
\color{OliveGreen}{\cmark}  &  \color{OliveGreen}{\cmark} & \color{red}{\xmark} \\
\hline
CG-PBR \cite{sengupta2019neural} & \color{OliveGreen}{\cmark} & \color{BurntOrange}{phong} & \color{red}{\xmark} & \color{BurntOrange}{shading} & \color{red}{\xmark} & \color{red}{\xmark} & \color{red}{\xmark} & \color{red}{\xmark} & \color{red}{\xmark} &  \color{OliveGreen}{\cmark} & \color{red}{\xmark} \\
\hline 
InvIndoor \cite{li2020inverse} & \color{OliveGreen}{\cmark} & \color{OliveGreen}{microfacet} & \color{red}{\xmark} & \color{OliveGreen}{envmap} & \color{red}{\xmark} & \color{red}{\xmark} & \color{red}{\xmark} & \color{red}{\xmark} & \color{OliveGreen}{\cmark} &  \color{OliveGreen}{\cmark} & \color{red}{\xmark} \\
\hline
3D-Future \cite{fu20203d} & \color{OliveGreen}{\cmark} & \color{red}{\xmark} & \color{red}{\xmark} & \color{red}{\xmark} & \color{red}{\xmark} & \color{OliveGreen}{\cmark} & \color{OliveGreen}{\cmark} & \color{OliveGreen}{\cmark} & \color{red}{\xmark}  &  \color{red}{\xmark} & \color{red}{\xmark} \\  
\hline
AI2-THOR \cite{kolve2017ai2} & \color{OliveGreen}{\cmark} & \color{red}{\cmark} & \color{OliveGreen}{\cmark} & \color{red}{\xmark} & \color{red}{\xmark} & \color{OliveGreen}{\cmark} & \color{OliveGreen}{\cmark} & \color{OliveGreen}{\cmark} & \color{red}{\xmark} & \color{red}{\xmark} & \color{OliveGreen}{\cmark} \\
\hline
Structure3D \cite{zheng2019structured3d} & \color{OliveGreen}{\cmark} & \color{red}{\xmark} & \color{red}{\xmark} & \color{orange}{shading} & \color{red}{\xmark} & \color{OliveGreen}{\cmark} &  \color{OliveGreen}{\cmark} & \color{red}{\xmark} & \color{red}{\xmark} & \color{OliveGreen}{\cmark} & \color{red}{\xmark}   \\
\hline
Hypersim \cite{roberts2020hypersim} & \color{OliveGreen}{\cmark} & \color{BurntOrange}{diffuse} & \color{red}{\xmark} & \color{BurntOrange}{highlight} & \color{red}{\xmark} & \color{OliveGreen}{\cmark} & \color{OliveGreen}{\cmark} & \color{red}{\xmark} &  \color{red}{\xmark} &  \color{OliveGreen}{\cmark} & \color{red}{\xmark} \\ 
\hline 
OpenRooms & \color{OliveGreen}{\cmark} & \color{OliveGreen}{microfacet} & \color{OliveGreen}{\cmark} & \color{OliveGreen}{envmap} & \color{OliveGreen}{\cmark} & \color{OliveGreen}{\cmark} & \color{OliveGreen}{\cmark} & \color{OliveGreen}{\cmark} & \color{OliveGreen}{\cmark} &  \color{OliveGreen}{\cmark} & \color{OliveGreen}{\cmark} \\
\hline
\end{tabular}
\vspace{-0.3cm}
\end{minipage}\hfill
\begin{minipage}[c]{0.23\textwidth}
\small
\caption{
OpenRooms is distinct in providing extensive ground truth for photorealism (especially material and lighting), with publicly available assets and tools. The tools in OpenRooms framework allow generating synthetic counterparts of real scenes, with high-quality ground truth.}
\label{tab:datasetComparison}
\vspace{-0.5cm}
\end{minipage}
\end{table*}

\section{Related Work}
\label{sec:related}
\vspace{-0.1cm}

\paragraph{Indoor scene datasets.} The importance of indoor scene reconstruction and understanding has led to a number of real datasets \cite{silberman2012indoor,dai2017scannet,chang17matterport3D,xia2018gibson,gibson_v1}. While they are by nature photorealistic, they only capture some scene information (usually images, geometry and semantic labels). However, we are interested in studying geometry, reflectance and illumination, where the latter two are particularly challenging to acquire in real datasets. Synthetic datasets provide an alternative \cite{mccormac2017scenenetrgbd,song2017semantic,li2018interiornet},
but prior ones are limited with respect to rendering arbitrary data \cite{li2018interiornet}, scene layout \cite{mccormac2017scenenetrgbd}, material  \cite{song2017semantic}, or baselines \cite{roberts2020hypersim}, as summarized in Table \ref{tab:datasetComparison}.

Several methods build 3D models for indoor scenes from a single image \cite{izadinia2017im2cad} or scans \cite{avetisyan2019scan2cad,avetisyan2020scenecad,cabral2014piecewise,chen2019floor}. 
However, our focus is beyond geometry, to assign real-world materials and lighting to create photorealistic scenes. To the best of our knowledge, the only existing dataset with complex materials and spatially-varying lighting annotations is from Li et al.~\cite{li2020inverse}, but is built on artist-created assets that are not publicly available \cite{song2017semantic}. We instead create photorealistic indoor scene datasets that start with 3D scans to provide high-quality ground truth for geometry, reflectance and lighting.

Several indoor virtual environments have also been proposed for robotics and embodied vision \cite{wu2018house3D,savva2017minos,xia2018gibson,gibson_v1,habitat19iccv,kolve2017ai2}. Our work is complementary, where our photorealistic ground truth and suite of tools could be used to enhance existing virtual environments and conduct new types of studies. In Sec.~\ref{sec:robotics}, we seek to motivate such adoption by illustrating integration with a physics engine and computing ground truth for friction coefficients.

\vspace{-0.5cm}
\paragraph{Inverse rendering for indoor scenes.} Indoor scene inverse rendering seeks to reconstruct geometry, reflectance and lighting from (in our case, monocular) RGB images. Estimating geometry, in the form of scene depth or surface normals, has been widely studied \cite{eigen2015depth,bansal2016marr,yao2018mvsnet,liu2018planenet}. 
Most scene material estimation methods either recognize material classes \cite{bell2015minc} or only reconstruct diffuse albedo \cite{li2018cgintrinsics,barron2013intrinsic,karsch-tog-14}. Scaling these methods to real-world images requires scene datasets with complex physically-based materials. 
Li et al.~\cite{li2020inverse} augment a proprietary dataset \cite{song2017semantic} with ground-truth SVBRDF annotations to train a physically-motivated network.
We demonstrate comparable inverse rendering performance using their network, but trained on OpenRooms, developed using publicly available assets.

Previous indoor scene lighting estimation methods only predict shading (which entangles geometry and lighting) \cite{li2018cgintrinsics}, require RGBD inputs \cite{barron2013intrinsic}, or rely on hand-crafted heuristics \cite{karsch-siga-11,karsch-tog-14}. More recently, deep network-based lighting estimation methods have shown great progress for estimating both global \cite{gardner2017indoor,gardner2019parametric} and spatially-varying lighting \cite{garon2019fast,Song_2019_CVPR,li2020inverse} from single RGB images. The latter set of methods largely rely on proprietary synthetic data to generate spatially-varying lighting annotations; we demonstrate comparable performance by training on our dataset.

\section{Building a Photorealistic Indoor Dataset}
\label{sec:dataset}
\vspace{-0.2cm}

We now describe our framework for building a synthetic dataset of complex indoor scenes. 
We demonstrate this using ScanNet, a large-scale repository of real indoor scans \cite{dai2017scannet}, but our work is also applicable to other datasets \cite{song2015rgbd,scenenn-3dv16}, as shown in the supplementary. 
We briefly describe the geometry creation, while focusing on our principal novelties of photorealistic material and lighting.

\begin{figure*}[t]
\centering
\includegraphics[width=\textwidth]{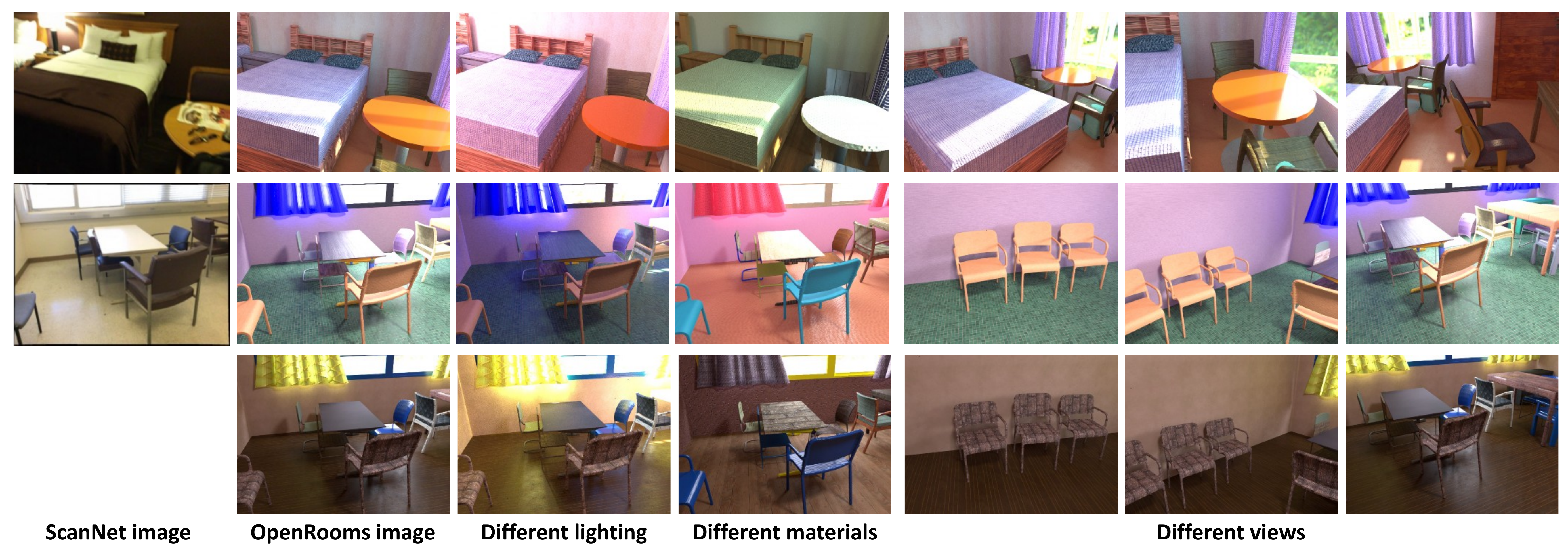}
\vspace{-0.3cm}
\caption{Images from ScanNet and our corresponding synthetic scene layouts rendered with different materials, different lighting, and different views selected by our algorithm. A video is included in the supplementary. The third row shows the same scene as the second one, but rendered with freely available Substance Share materials \cite{substance} instead of the public but non-free Adobe Stock materials \cite{adobestock}.}
\vspace{-0.4cm}
\label{fig:dataset}
\end{figure*}

\begin{figure}[t]
\centering
\includegraphics[width=0.95\columnwidth]{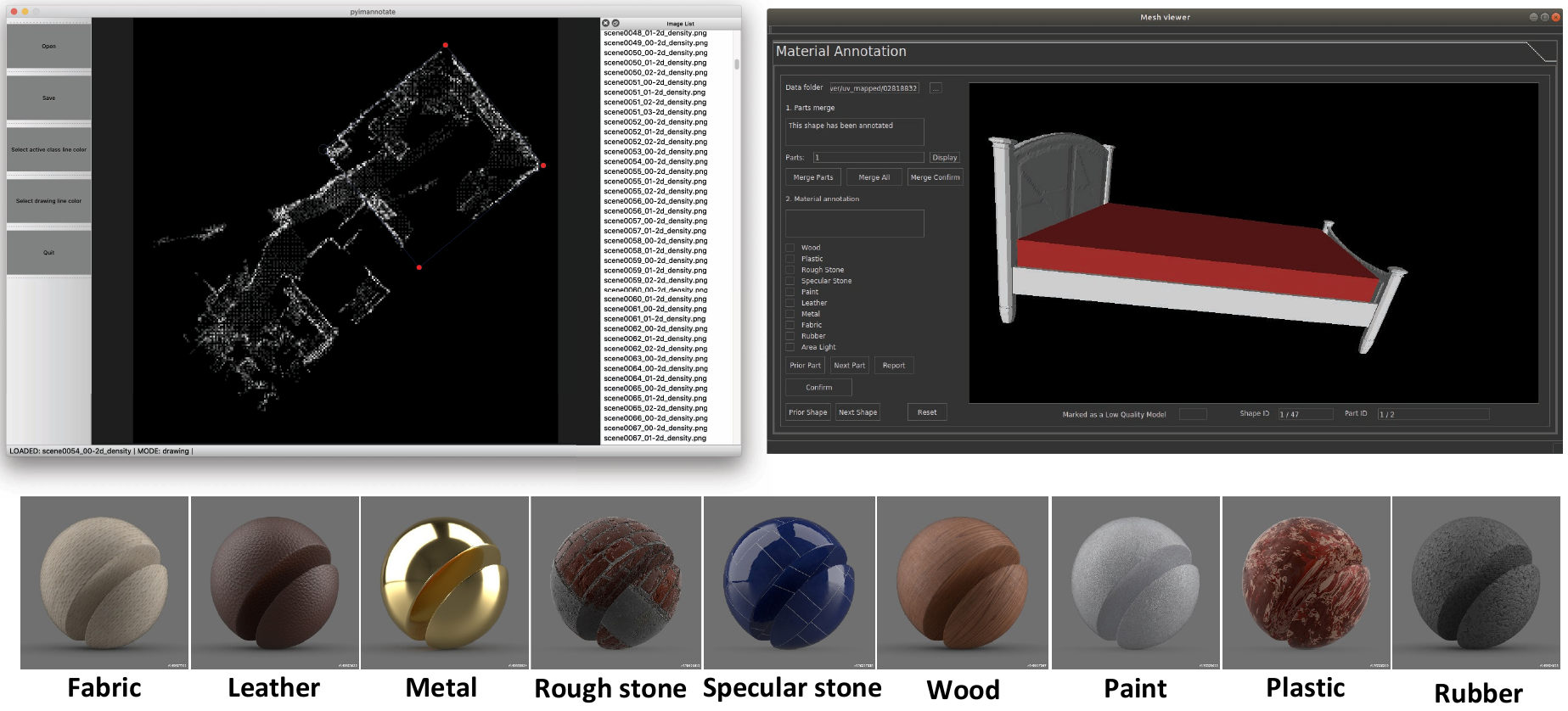}
\vspace{-0.3cm}
\caption{UIs for annotating room layout (Left top) and material category (Right top). (Bottom) Material examples from each category. Please zoom in for better visualization. }
\label{fig:uimat}
\vspace{-0.4cm}
\end{figure}

\vspace{-0.1cm}
\subsection{Creating CAD Models from 3D Scans}
\vspace{-0.1cm}

While recent methods such as \cite{avetisyan2020scenecad} are possible alternatives, we demonstrate our dataset creation example utilizing existing labels in ScanNet and initial CAD alignment \cite{avetisyan2019scan2cad} to create the ground truth geometry robustly. 

\vspace{-0.4cm}
\paragraph{Reconstructing the room layout}
We fuse the depth maps from different views of a scene to obtain a single point cloud. We design a UI for fast layout annotation (Fig.~\ref{fig:uimat}), which projects the 3D point cloud to the floor plane and a polygon may be selected for the layout. While the annotation needs less than a minute per scene, we also train a Floor-SP network \cite{chen2019floor} on these annotations that users may employ for their own scenes (shown in the supplementary). Next we use RANSAC to determine the horizontal floor plane. Since ScanNet views generally do not cover the ceiling, we assign a constant room height of 3 meters.

\vspace{-0.4cm}
\paragraph{Windows and doors} Special consideration is needed for doors and windows as they are important illuminants in indoor scenes. We project the 3D points labeled as doors and windows to the closest wall, then divide the wall into segments and merge connected segments with sufficient number of points, to which a ShapeNet CAD model is assigned.

\vspace{-0.4cm}
\paragraph{Consistent furniture placement} 
We use initial poses from Scan2CAD \cite{avetisyan2019scan2cad} to align CAD models with furniture instances. We do not require appearances to closely match the input images, but generate plausible layouts and shapes with as much automation as possible. Our tool automatically moves bounding boxes for  furniture perpendicular to the floors and walls to resolve floating objects and intersections. 
Such geometric consistency is important since our dataset may also be used for tasks such as navigation.

\begin{figure*}[!!t]
\centering
\includegraphics[width=\linewidth]{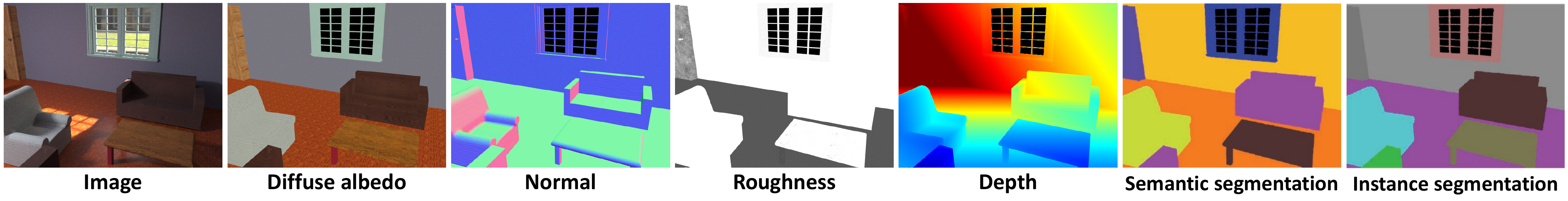}
\vspace{-0.4cm}
\caption{One of our rendered images with ground-truth geometry, spatially-varying material and segmentation labels. }
\label{fig:dataset_gt}
\vspace{-0.35cm}
\end{figure*}

\vspace{-0.4cm}
\paragraph{Semantic labels} 
Given our geometry ground truth, it is straightforward to obtain labels for semantic and instance segmentation based on PartNet annotations, as shown in Fig.~\ref{fig:dataset_gt}. We demonstate in experiments that our labels can be used to train single and multi-task deep networks.

\vspace{-0.1cm}
\subsection{Assigning Complex Materials to Indoor Scenes}
\vspace{-0.2cm}
One of the major contributions of our dataset is ground-truth annotation of complex material parameters for indoor scenes.  Previous works typically provide material annotations as simple diffuse or Phong reflectance \cite{song2017semantic,shi2017learning}, while we provide a physically-based microfacet SVBRDF.

\vspace{-0.4cm}
\paragraph{Assigning materials to ShapeNet} 
Many ShapeNet CAD models do not have texture coordinates, so we use Blender's \cite{blender} cube projection UV mapping to compute texture coordinates for them automatically. Inspired by Photoshape \cite{park2019photoshape}, we split CAD models into semantically meaningful parts and assign a material to each part. While Photoshape does this for only chairs, we do so for all furniture types in indoor scenes, using the semantically meaningful part segmentation of 24 categories of models provided by PartNet \cite{mo2019partnet}. 

\vspace{-0.4cm}
\paragraph{Material annotation UI} 
We design a custom UI tool to annotate material category for each part, as shown in Fig.~\ref{fig:uimat}. It allows merging over-segmented parts which should be assigned the same material.
To allow material annotation, we group 1,078 SVBRDFs into 9 categories based on their appearances, similar to \cite{li2018material,li2020inverse}, as shown in Fig.~\ref{fig:uimat}. Annotators label a material category for each part, with a specific material sampled randomly from the category. While we do not pursue mimicking input appearances, we do seek that photorealism and semantics be respected in the dataset.
Experiments show that our dataset created following the above choices  enables state-of-the-art inverse rendering performances.  Note our distinction from domain randomization, since arbitrary choices for material and lighting might not allow generalization on real scenes for extremely ill-posed problems like material and lighting estimation. Our tools and the annotations will be released for future research.

\subsection{Ground Truth Lighting for Indoor Scenes}
\vspace{-0.2cm}
Lighting plays one of the most important roles in image formation. However, prior datasets usually only provide diffuse shading as their lighting representation \cite{li2018cgintrinsics,zhang2016physically}. Recent work provides per-pixel environment maps by rendering the incoming radiance at every surface point in the camera frustum \cite{li2020inverse}, which allows modeling shadows and specular highlights, but not the complex interactions among global light sources, scene geometry, materials and local lighting. On the contrary, OpenRooms provides extra supervision for visible and invisible light sources, the contribution of each individual light source to the local lighting, direct and indirect lighting, as well as visibility. Such rich supervision may help better understand the complex light transport in indoor scenes and enable new applications such as editing of light sources and dynamic scenes. 

\begin{figure}[!!t]
\centering
\includegraphics[width=0.93\columnwidth]{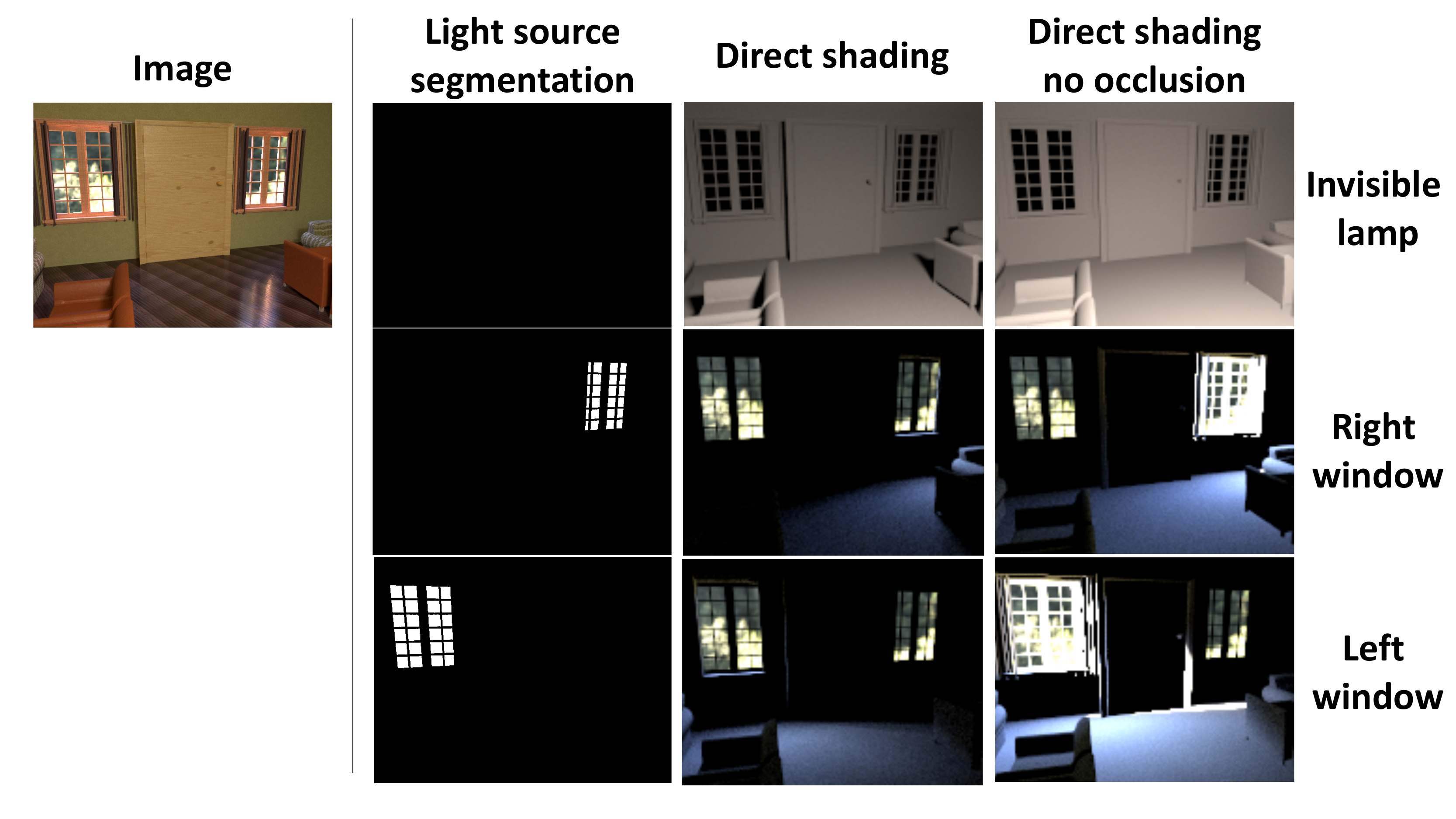}
\vspace{-0.4cm}
\caption{Our ground-truth light source annotations. From left to right: input and for each light source, its instance segmentation, and direct shading with and without occlusion. Our annotations reveal rich information about light transport in indoor scenes.}
\label{fig:dataset-lightSource}
\vspace{-0.3cm}
\end{figure}

\vspace{-0.4cm}
\paragraph{Light sources}
We model two types of light sources in OpenRooms---windows and lamps---and we provide ground-truth annotations for them. The annotations include instance segmentation masks for visible light sources and a consistent parameterized representation for both visible and invisible light sources. 
More specifically, for each window, we model its geometry using a rectangular plane and the lighting coming through the window using an environment map rendered at its center. 
We represent each lamp as a 3D bounding box following the standard area light model. 
We visualize our light source annotations in Figure \ref{fig:dataset-lightSource}. 
Our light source representation has clear physical meaning and can model the full physics of image formation in indoor scenes. 

\vspace{-0.4cm} 
\paragraph{Light source colors}
For environment maps, we use 414 high-resolution HDR panoramas of natural outdoor scenes, from \cite{holdgeoffroy19sky} and \cite{HDRI}. For indoor lamps, unlike previous synthetic datasets that randomly sample the spectrum of area lights \cite{li2020inverse,zhang2016physically,li2018cgintrinsics}, we follow a physically plausible black-body model to determine the spectrum of the light source by its temperature, chosen between 4000K to 8000K.

\begin{figure}[!!t]
\centering
\includegraphics[width=0.95\columnwidth]{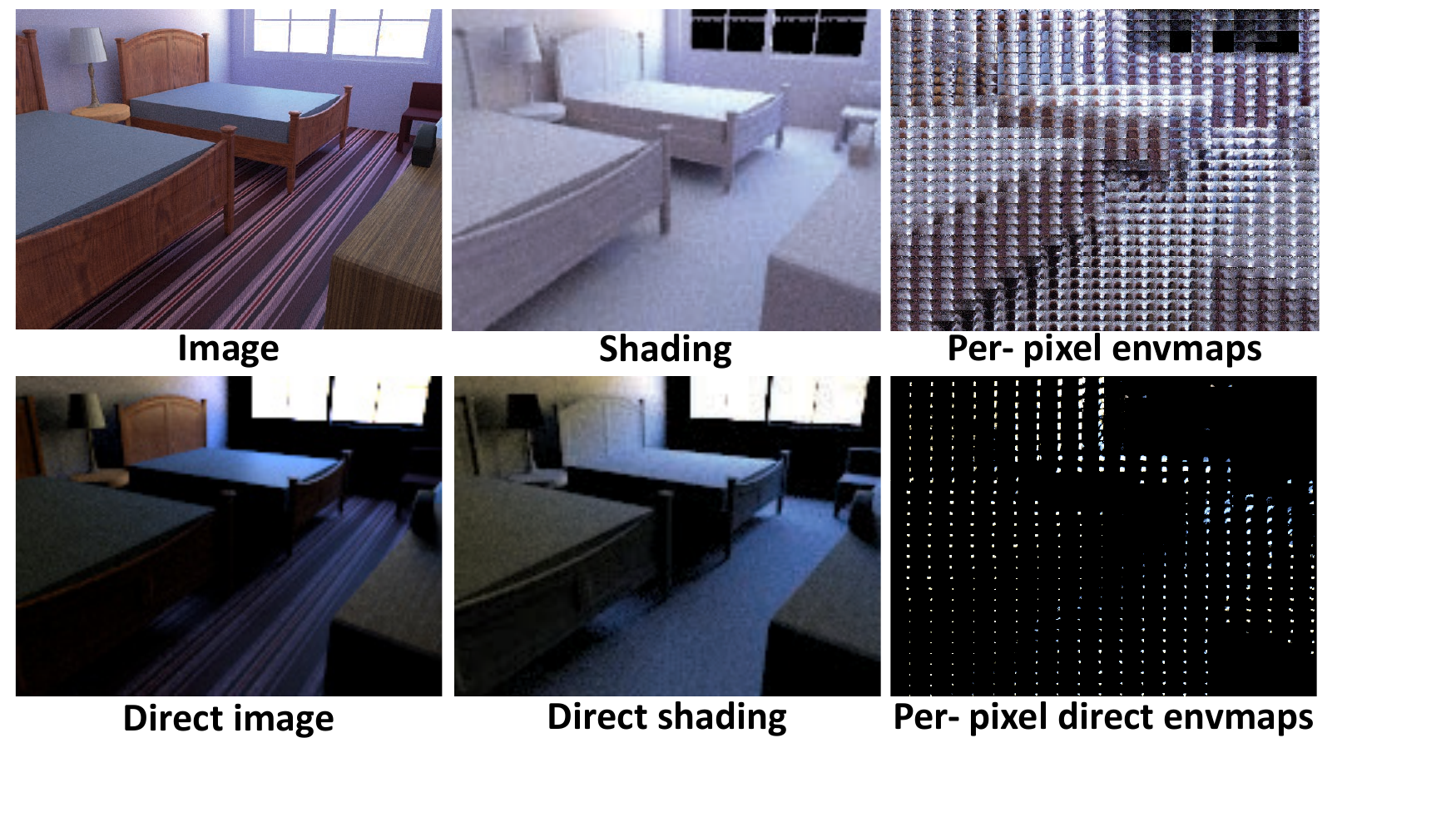}
\vspace{-0.4cm}
\caption{We provide various types of supervision for lighting analysis of indoor scenes, including per-pixel environment maps with only direct illumination, or including indirect illumination.}
\label{fig:dataset-light}
\vspace{-0.4cm}
\end{figure}

\vspace{-0.4cm}
\paragraph{Per-pixel lighting}
Additionally, as in prior works \cite{li2020inverse,li2018cgintrinsics,zhang2016physically}, we render per-pixel environment maps and shading as a spatially-varying lighting representation. However, we render both with direct, as well as combined direct and indirect illumination. This will help to separately analyze the direct contribution from light sources and indirect reflections from the indoor scene. We visualize an example in Figure \ref{fig:dataset-light}. 

\vspace{-0.4cm} 
\paragraph{Per-light direct shading and visibility}
In order to understand complex light transport in indoor scenes, we also provide the separate contribution of every individual light source and its visibility map. For each image, we render the direct shading of each light source, with and without considering the occlusion term, by turning on only that particular light source. The visibility map can be computed as the ratio of the two direct shading images. We visualize these annotations in Figures \ref{fig:dataset-lightSource} and \ref{fig:dataset-light}. These will allow new challenging light editing tasks not possible with prior datasets, such as turning on and off a light or opening a window.

\subsection{Rendering with a Physically-based Renderer} 
\vspace{-0.2cm}

To minimize the domain gap between synthetic and real data, we modify the physically-based GPU-accelerated renderer from our prior work \cite{li2020inverse} to support ground-truth per-light contribution and fast rendering of per-pixel environment map. Our renderer models complex light transport up to 7 bounces of inter-reflection.

\vspace{-0.4cm}
\paragraph{View selection}
ScanNet provides the camera pose of each RGBD image. However, their distribution is biased towards views close to the scene geometry, to optimize scanning. On the contrary, we prefer views covering larger regions, matching typical human viewing conditions. To achieve this, we first sample different views along the wall, facing the center of the room. For each view, we render its depth and normal maps. Let $d_{p}$ and $\mathbf{\hat{n}}_p$ be the depth and normal of pixel $p$, $\mathbf{Grad}(\mathbf{\hat{n}}_{p})$ be the sum of absolute gradients of the normal in the three channels. We choose the view based on computing a score defined as
\begin{equation}
\sum_{p\in\mathcal{P}} \mathbf{Grad}(\mathbf{\hat{n}}_p) + 0.3 \sum_{p \in \mathcal{P}} \log(d_p + 1).
\end{equation}
Views with higher scores are used to create the dataset. An example of our view selection results is shown in Figure \ref{fig:dataset} (bottom right). Details are included in the supplementary. 

\vspace{-0.4cm}
\paragraph{Other renderers} While our renderer will be publicly released, our assets (geometry, material maps, lights) are in a standard graphics format that could be used in other rendering environments. For example, common real-time rasterization engines like Unity or Unreal could be used for applications (such as robotics) which prefer real-time performance and do not require fully accurate global illumination. Furthermore, our per-pixel spatially-varying lighting maps could be used as high-quality precomputed lighting probes for photorealistic real-time rendering \cite{mcguire2017lightfield}.

\begin{figure*}
\centering
\includegraphics[width=\linewidth]{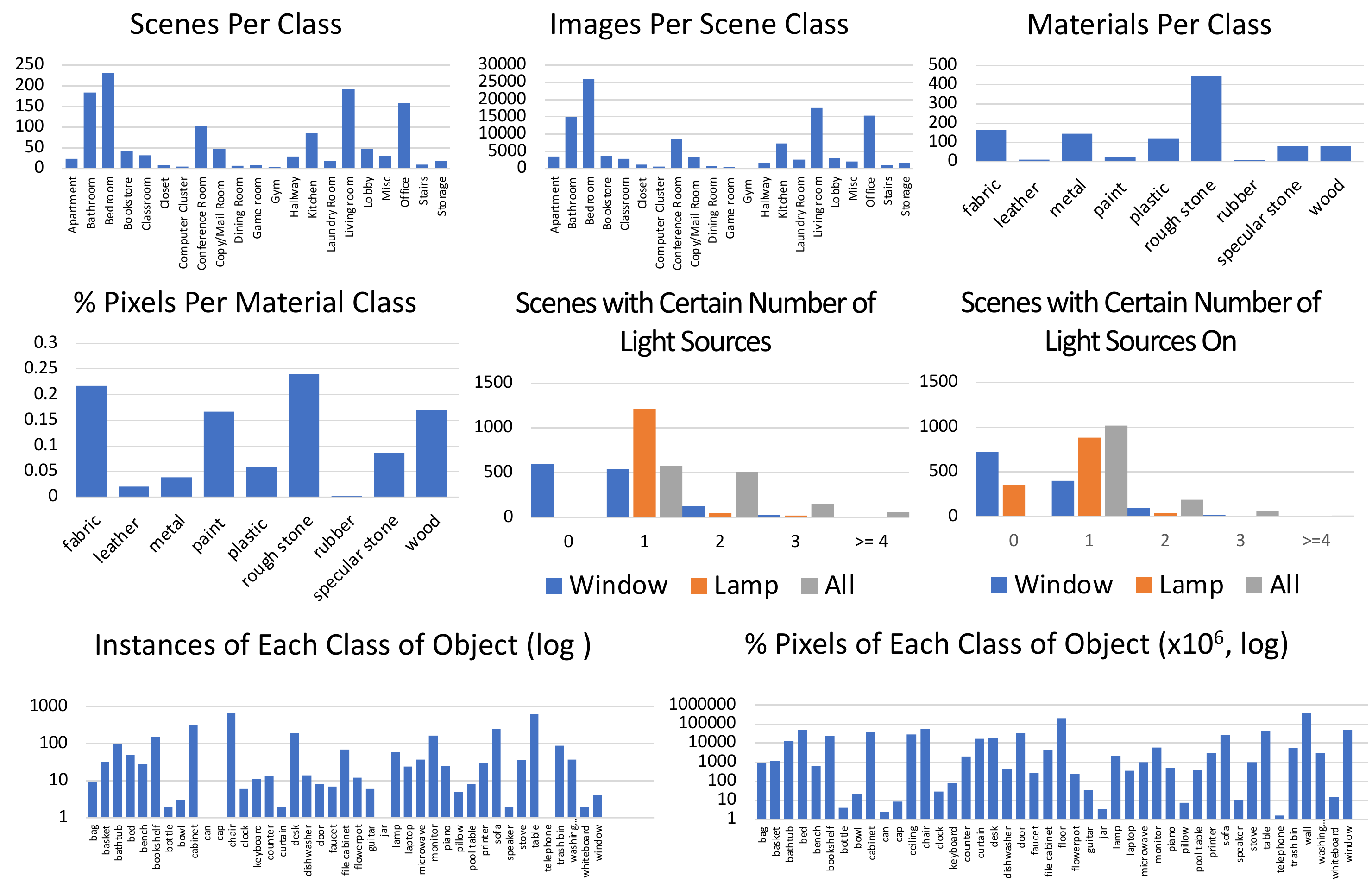}
\vspace{-0.3cm}
\caption{Dataset statistics for scene categories, images, materials, lighting and semantic labels (please zoom for viewing).}
\label{fig:dataset-dists}
\vspace{-0.4cm}
\end{figure*}

\subsection{OpenRooms Dataset Statistics}

\vspace{-0.1cm}
\paragraph{Scene, image, semantic label distribution} We pick 1,287 of the 1,506 ScanNet scenes to instantiate our dataset, discarding those which cover very small portions of rooms. We randomly choose 1,178 scenes for training and 109 scenes for validation. 
For each scene, we choose views using our view selection method. For each rendered image, we render two others with different materials and lighting, as shown in Fig.~\ref{fig:dataset} (bottom-left). We render 118,233 HDR images at $480\times 640$ resolution, with 108,159 in the training set and 10,074 in the validation set. 
We render semantic labels of all 44 classes of CAD models in OpenRooms. 
The distributions of scene categories and images, number of objects per class and the percentage of pixels per class are summarized in Figure \ref{fig:dataset-dists}. Note that the class distribution follows that of real scans in ScanNet indoor scenes.

\vspace{-0.4cm}
\paragraph{Material distribution} 
We use 1,075 SVBRDFs from \cite{adobestock} to build OpenRooms, corresponding to the 9 categories shown in Fig.~\ref{fig:uimat}.
The number of materials per-category and their pixel distributions are summarized in Fig.~\ref{fig:dataset-dists}.

\vspace{-0.4cm}
\paragraph{Lighting distribution}
Figure~\ref{fig:dataset-dists} shows the distribution of the two types of light sources (windows and lamps). Each image has at least one light source ``on'' for rendering. For all the 118K images, we render spatially-varying environment maps and shading, with direct illumination only and with combined direct and indirect illumination. Moreover, we provide a parameterized representation for every visible and invisible light source, as well as render their individual direct shading contribution and visibility map. Compared to all prior works, OpenRooms provides significantly more extensive and detailed supervision for complex lighting, which may allow new applications such as light source detection and editing.

\vspace{-0.4cm}
\paragraph{Asset cost}
Almost all the assets used for creating our dataset are publicly available and free for research use. The only non-free (but also publicly available) assets are the raw material maps from Adobe Stock \cite{adobestock} that cost less than US\$500, while the material parameters annotated with our scenes are freely available. Note that photorealistic appearances may also be achieved using our tools with freely available materials, such as Substance Share \cite{substance} in Fig.~\ref{fig:dataset}.

\vspace{-0.4cm}
\paragraph{Dataset creation time}
It takes 30s to annotate one scene layout and 1 minute to label materials for one object, leading to 64 hours for labeling the whole dataset, which was accomplished by students with knowledge of computer vision. Almost all rendering time is spent to render images and spatially-varying per-pixel environment maps, which takes 600s and 100s per image, respectively, for our custom renderer on a single 2080Ti GPU. In principle, we can render the dataset in 1 month using 40 GPUs.

\section{Applications}
\vspace{-0.1cm}
\subsection{Inverse Rendering}
\label{sec:experiments}
\vspace{-0.1cm}

We verify the effectiveness for inverse rendering by testing networks trained on our dataset on various benchmarks, where both quantitative and qualitative results show good generalization to real images. We use a state-of-the-art network architecture for inverse rendering in indoor scenes that handles spatially-varying material and lighting \cite{li2020inverse}. Please refer to the supplementary material for more details. 

\begin{figure}
\centering
\includegraphics[width=\columnwidth]{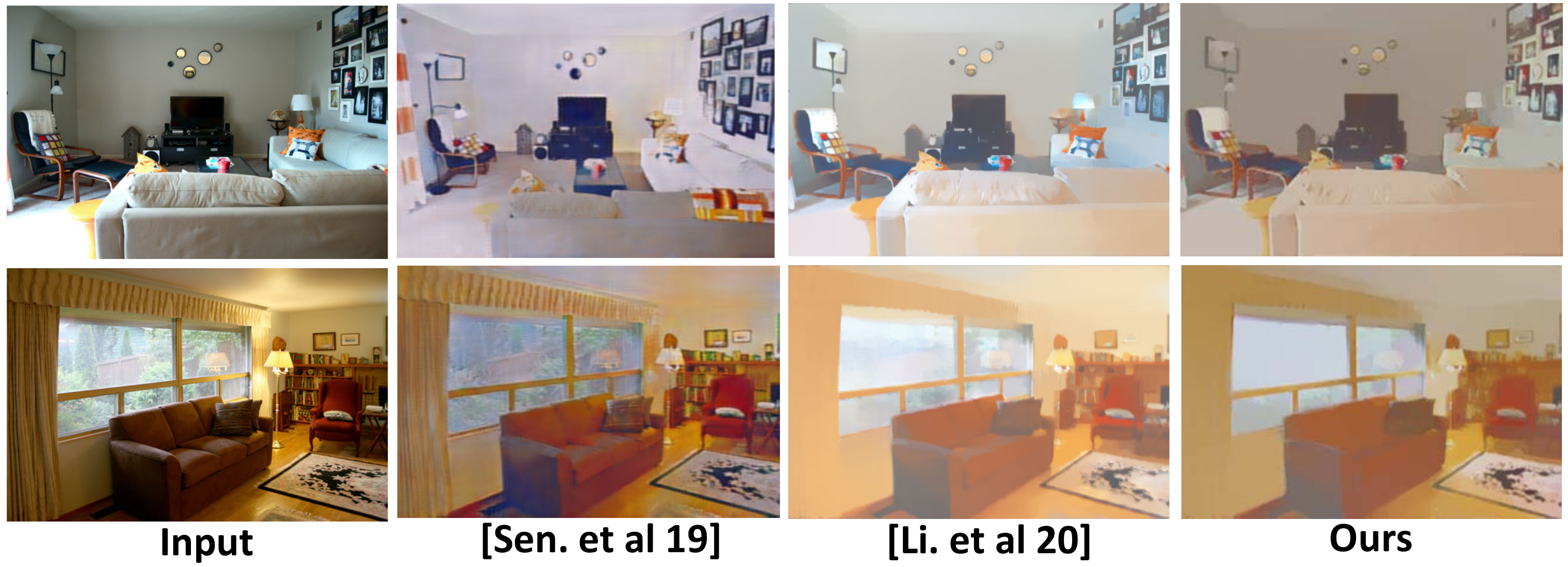}
\vspace{-0.3cm}
\caption{Comparisons with previous state-of-the-art on intrinsic decomposition (albedo prediction shown). }
\label{fig:intrinsic}
\end{figure}

\begin{figure*}
\centering
\includegraphics[width=\textwidth]{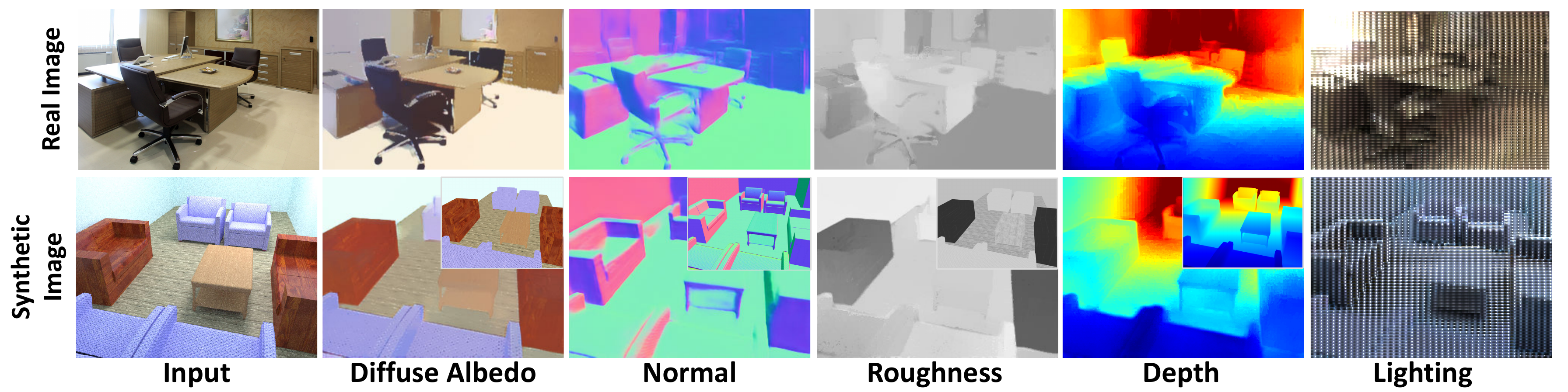}
\vspace{-0.3cm}
\caption{Inverse rendering results on a real example and a synthetic example. The insets in the bottom row are the ground truth.}
\label{fig:real}
\vspace{-0.4cm}
\end{figure*}

\begin{table}
\footnotesize
  \begin{center}
    \begin{minipage}{0.6\linewidth}
        \begin{tabular}{|c|c|c|}
        \hline
         & Training & WHDR$\downarrow$ \\
        \hline
        \textbf{Ours}  & Ours + IIW & 16.4\\
        \hline
        Li18\cite{li2018cgintrinsics} & CGI + IIW & 17.5 \\
        \hline
        Sen.19\cite{sengupta2019neural} & CGP + IIW & 16.7\\
        \hline
        Li20\cite{li2020inverse} & CGM + IIW & 15.9\\
        \hline
        \end{tabular} 
    \end{minipage}\hfill
    \begin{minipage}{0.38\linewidth}
      \vspace{-0.5cm}
      \caption{
       Intrinsic decomposition on IIW \cite{bell2014intrinsic}.
      }
      \label{tab:IIW}
    \end{minipage}
  \end{center}
  \vspace{-0.5cm}
\end{table}

\begin{table}
\footnotesize
  \begin{center}
    \begin{minipage}{0.7\linewidth}
        \addtolength{\tabcolsep}{-1pt}    
        \begin{tabular}{|c|c|c|c|}
        \hline 
        Method & Mean($^{\circ}$)$\downarrow$ & Med.($^{\circ}$)$\downarrow$ & Depth$\downarrow$  \\ 
        \hline
        \textbf{Ours}  & 25.3 & 18.0 & 0.171   \\
        \hline
        Li20\cite{li2020inverse} & 24.1 & 17.3 & 0.184 \\
        \hline
        Sen.19\cite{sengupta2019neural} & 21.1 & 16.9 & -- \\
        \hline
        Zhang17\cite{zhang2016physically} & 21.7 & 14.8 & -- \\
        \hline
        \end{tabular}
        \addtolength{\tabcolsep}{1pt}    
    \end{minipage}\hfill
    \begin{minipage}{0.28\linewidth}
      \caption{
       Normal and depth predictions on NYU dataset \cite{silberman2012indoor}.
      }
      \label{tab:NYU}
    \end{minipage}
  \end{center}
  \vspace{-0.5cm}
\end{table}

\vspace{-0.2cm}
\paragraph{Intrinsic decomposition.} We compare our intrinsic decomposition results with 3 previous approaches. The qualitative comparison is shown in Fig.~\ref{fig:intrinsic} while quantitative results are in Table \ref{tab:IIW}, which are comparable to prior state-of-the-art based on artist-created SUNCG dataset \cite{song2017semantic}.

\vspace{-0.2cm}
\paragraph{Depth and normal estimation.} We evaluate the normal and depth estimation on the NYU dataset. The quantitative evaluation is in Table \ref{tab:NYU}. We perform slightly worse than Li et al.'s dataset, possibly because their SUNCG-based dataset has more diverse and complex geometry compared to our ShapeNet-based furnitures.

\begin{figure}
\centering
\includegraphics[width=\columnwidth]{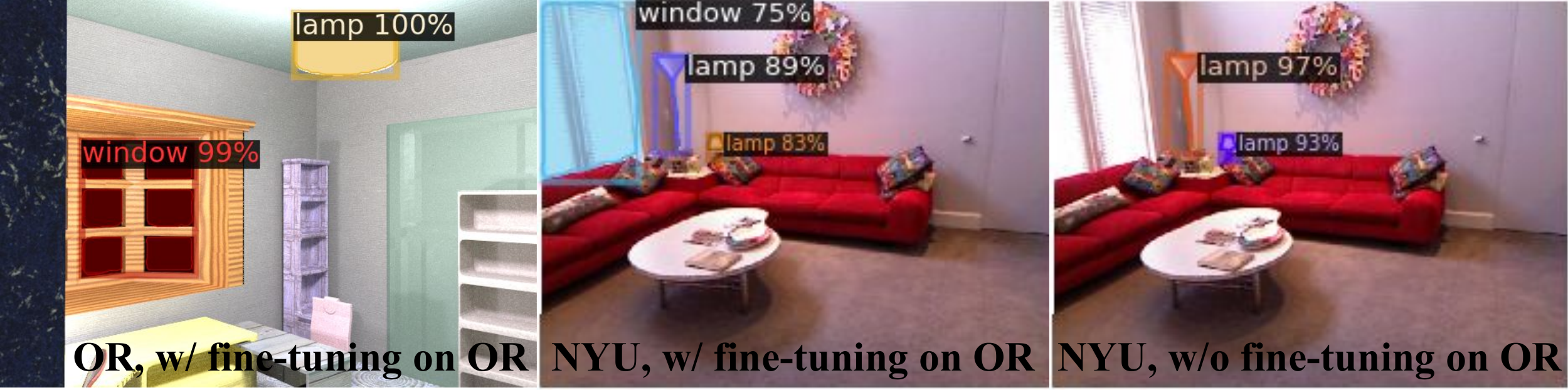}
\vspace{-0.5cm}
\caption{Light source detection on OpenRooms (OR) and NYUv2~\cite{silberman2012indoor}. Windows are better detected with OR training.}
\label{fig:instance}
\vspace{-0.3cm}
\end{figure}

\vspace{-0.2cm}
\paragraph{Light source detection}
We use a ResNeXt101~\cite{xie2017aggregated} and FPN \cite{8099589} pretrained model from Detectron2~\cite{wu2019detectron2} to train an instance segmentation network for light source detection (windows and lamps). We evaluate on OpenRooms and NYUv2~\cite{silberman2012indoor}. As shown in Tab.~\ref{tab:instance} and Fig.~\ref{fig:instance}, training on OpenRooms boosts accuracy on NYUv2 testing by around $5\%$, for both bounding box regression and segmentation.

\begin{table}
\scriptsize
\addtolength{\tabcolsep}{-1.1pt}  
\begin{center}
\begin{tabular}{ |c|cc|cc|cc| }
  \hline
    \multicolumn{1}{|c|}{ Test on}&
    \multicolumn{2}{|c|}{ OpenRoom } &
    \multicolumn{4}{|c|}{NYU2 } 
   \\
 \hline
    \multicolumn{1}{|c|}{Train on OR/NYU2}&
    \multicolumn{2}{|c|}{Yes/ No } &
    \multicolumn{2}{|c|}{No/ Yes }&
    \multicolumn{2}{|c|}{Yes/ Yes } 
   \\
 \hline
    &  bbox &  seg &  bbox &  seg &bbox &  seg \\

   \hline
    AP(0.5:0.95)&  80.2 &  70.1 &  17.1 &  15.3 & 23.5 &  21.6 \\
    AP-windows&  85.8 &  63.2 &  11.9 &  12.7 &20.5 &  20.6 \\
    AP-lamp&  74.7 &  76.9 &  22.2 &  18.0 &26.6 &  22.7 \\
 \hline
\end{tabular}
\vspace{-0.2cm}
\caption{Bounding box regression and mask AP on OpenRooms and NYU2~\cite{silberman2012indoor} for light source (windows and lamps) detection.
}
\label{tab:instance}
\end{center}
\addtolength{\tabcolsep}{1.1pt}  
\vspace{-0.5cm}
\end{table}

\begin{figure}
\centering
\includegraphics[width=\columnwidth]{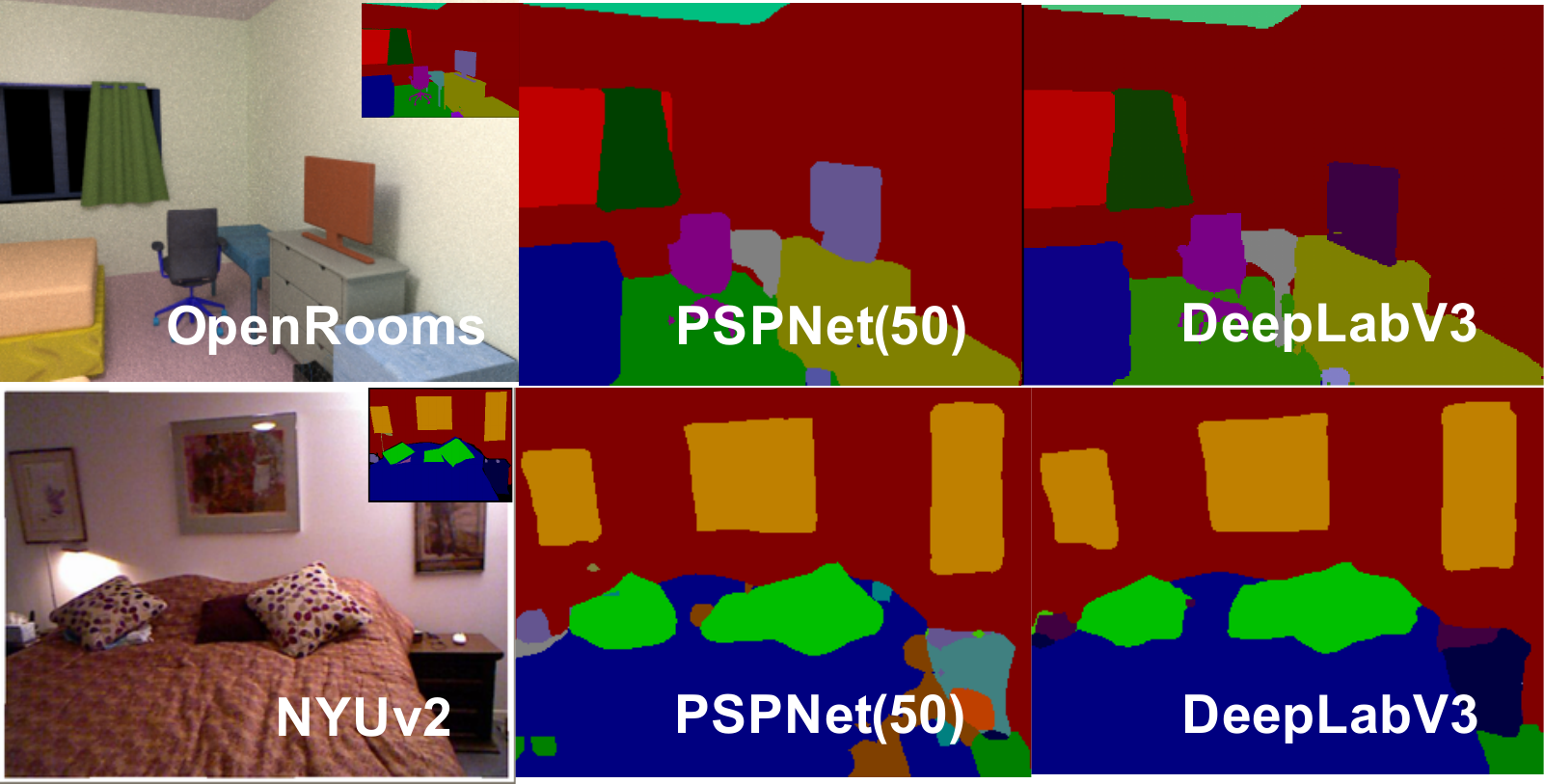}
\vspace{-3mm}
\caption{Semantic segmentation on OpenRooms and NYUv2~\cite{silberman2012indoor} using PSPNet(50)~\cite{zhao2017pyramid} and DeepLabV3~\cite{chen2017rethinking}.}
\label{fig:segmentation}
\vspace{-0.2cm}
\end{figure}

\begin{figure}
\centering
\includegraphics[width=\columnwidth]{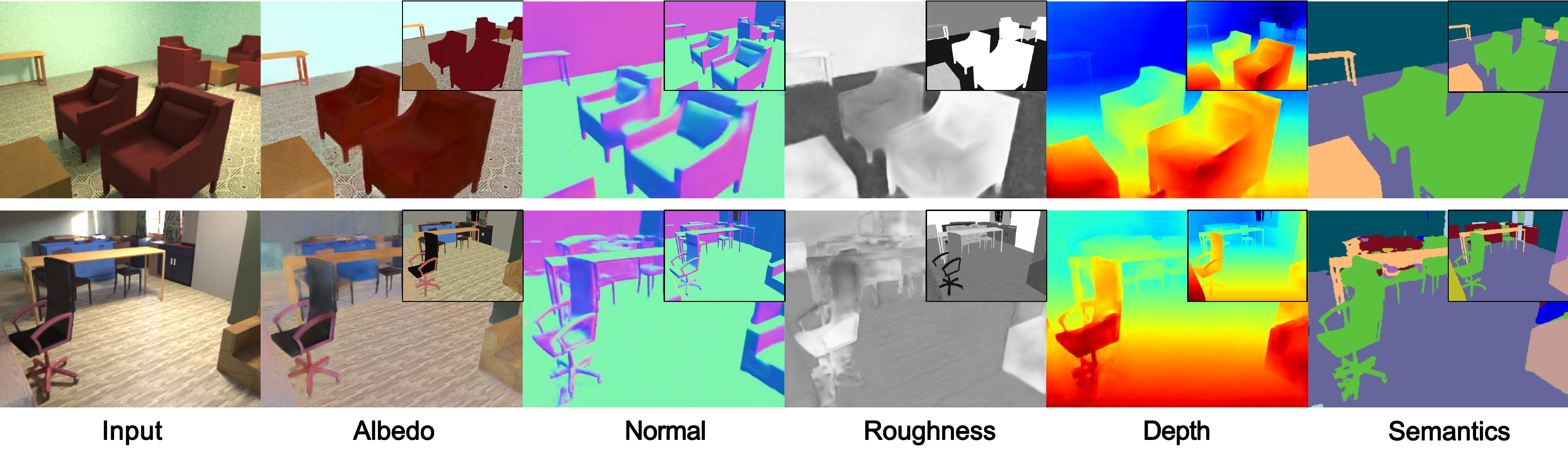}
\vspace{-0.6cm}
\caption{Multi-task estimation on OpenRooms.}
\label{fig:multitask}
\vspace{-0.4cm}
\end{figure}

\begin{figure}[t]
\centering
\includegraphics[width=\columnwidth]{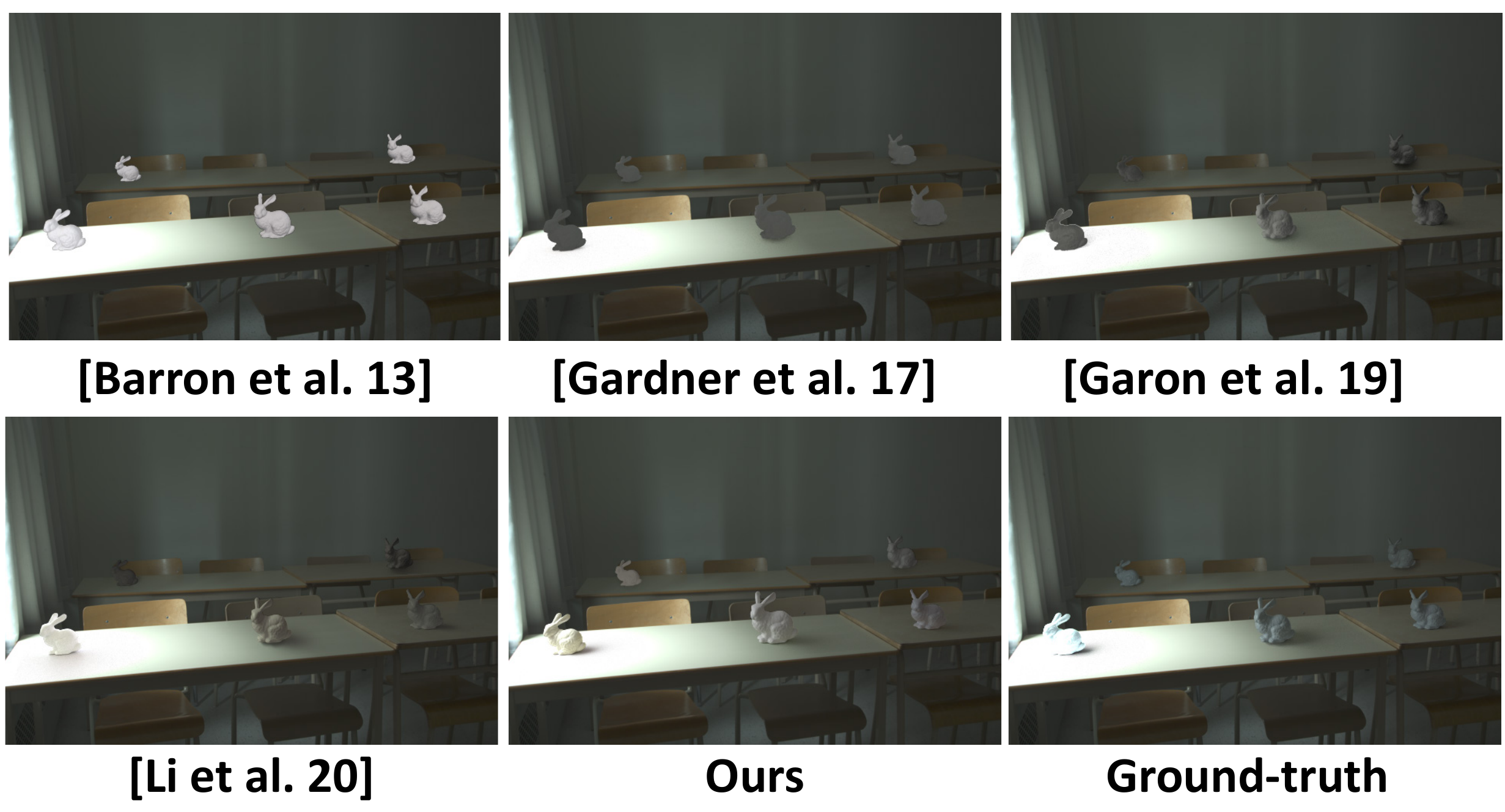}
\vspace{-0.6cm}
\caption{Object insertion on a real benchmark dataset \cite{garon2019fast}. Our dataset leads to photorealistic insertion results comparable to state-of-the-art \cite{li2020inverse}\cite{garon2019fast}. Please zoom in for more details. }
\label{fig:objectInsertion}
\vspace{-0.2cm}
\end{figure}

\begin{figure}[!!t]
\centering
\includegraphics[width=\columnwidth]{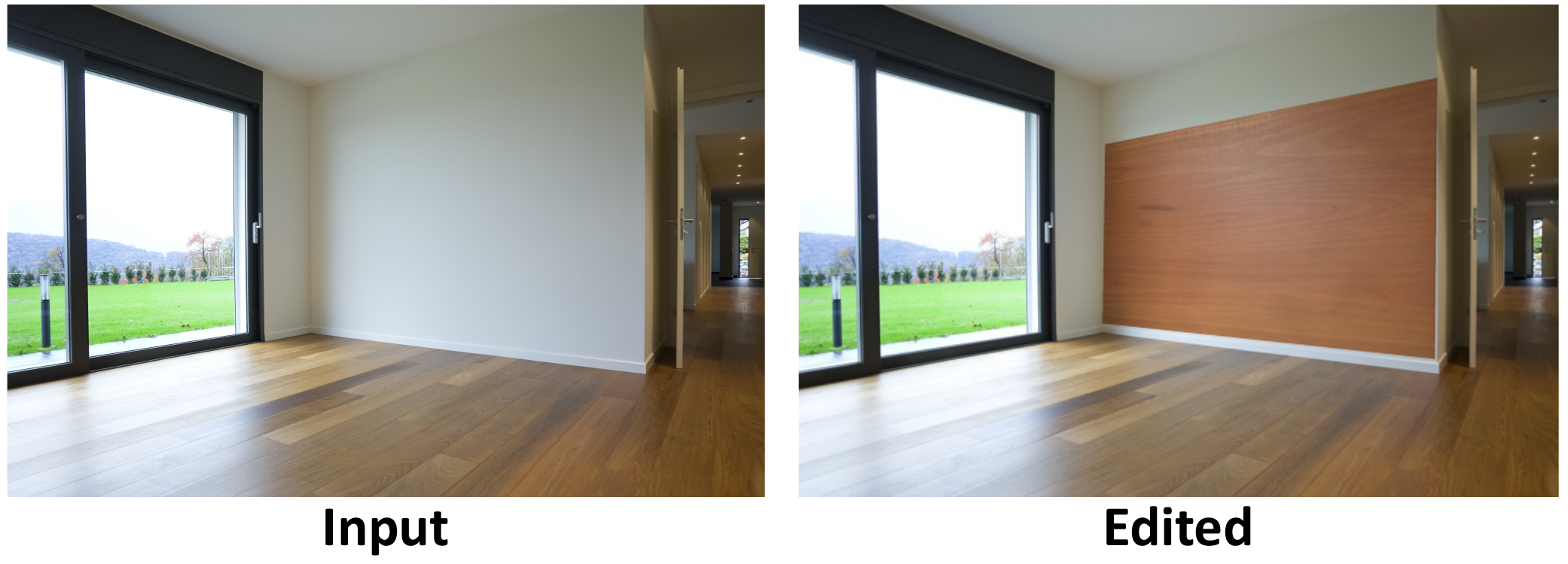}
\vspace{-6mm}
\caption{Material editing in real images. Note that the network trained on our dataset handles specular effects and spatially-varying lighting well.}
\label{fig:matReplacement}
\vspace{-0.2cm}
\end{figure}

\begin{figure*}[!!t]
\centering
    \includegraphics[width=\textwidth]{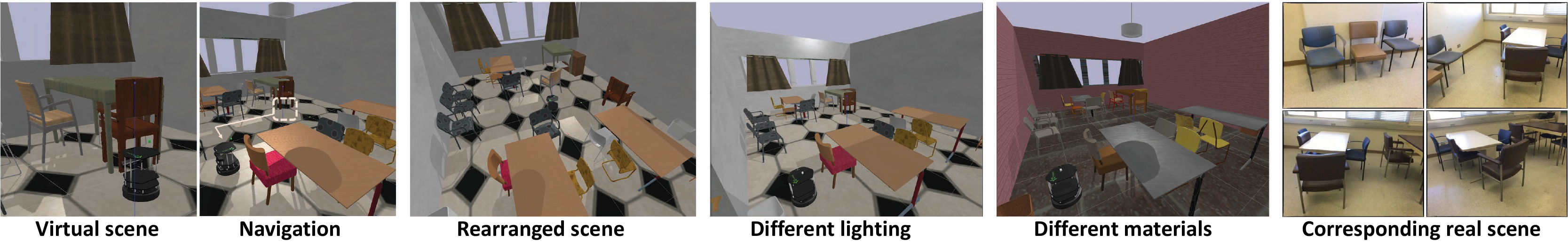}
\vspace{-0.3cm}
\caption{OpenRooms is integrated with a physics engine to create virtual scenes for robotics, potentially enabling studies for navigation and rearrangement across varying lighting and material, with possible correspondence to real scenes.}
\label{fig:robotics}
\vspace{-0.4cm}
\end{figure*} 

\begin{figure}[!!t]
\centering
\includegraphics[width=\columnwidth]{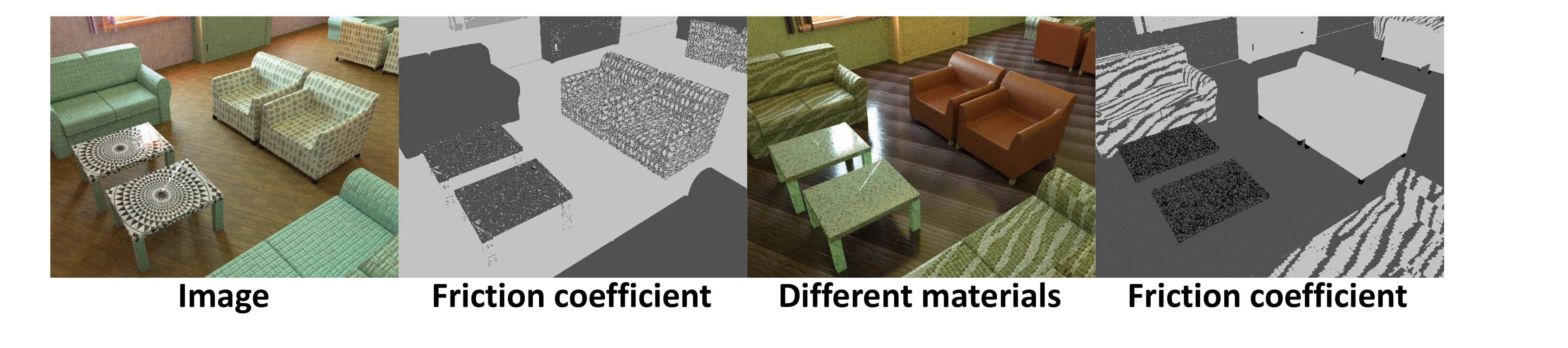}
\vspace{-0.7cm}
\caption{Ground-truth friction coefficients for the same scene with different materials. Specular materials tend to have lower coefficients of friction (darker). }
\label{fig:friction}
\vspace{-0.4cm}
\end{figure}

\begin{figure}[!!t]
\centering
\includegraphics[width=\columnwidth]{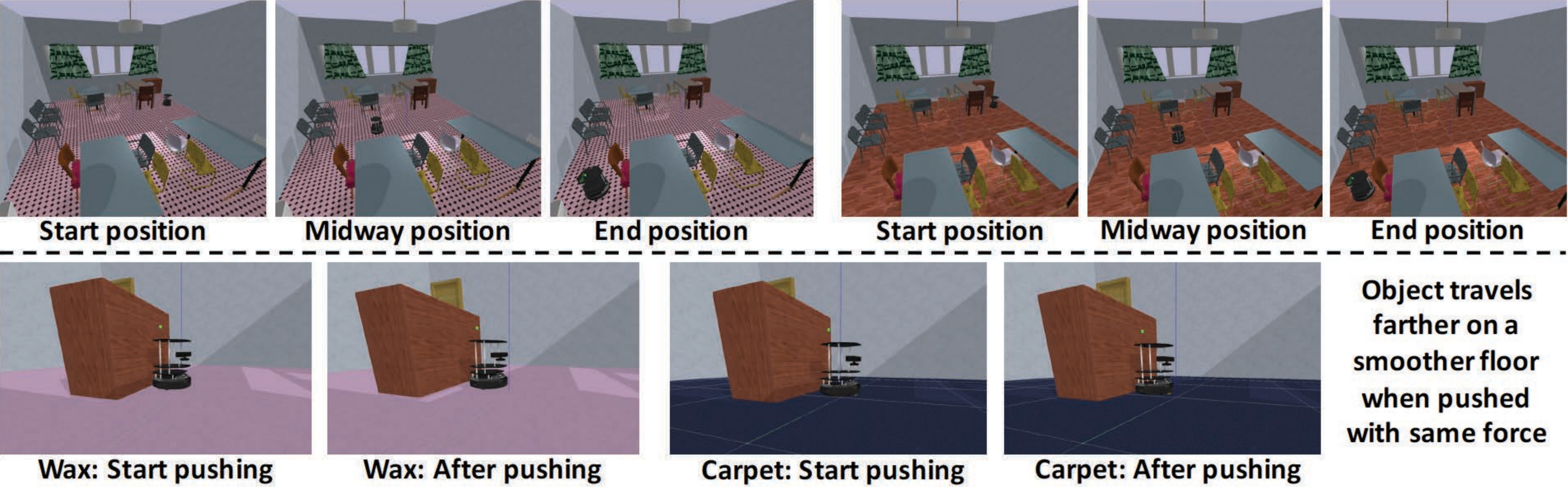}
\vspace{-0.6cm}
\caption{OpenRooms enables novel studies in navigation and rearrangement with material and lighting variations.}
\label{fig:robotlearningtasks}
\vspace{-0.4cm}
\end{figure}

\vspace{-0.4cm}
\paragraph{Per-pixel lighting estimation}
The above network also predicts per-pixel spatially-varying lighting, with qualitative results shown in Fig.~\ref{fig:real} and quantitative results in supplementary. Note that we also provide ground truth for per-pixel direct lighting, shading and visibility, which are not predicted by our network but may be useful for studies in light transport, editing and augmented reality.

\vspace{-0.4cm}
\paragraph{Semantic segmentation}
We use DeepLabV3~\cite{chen2017rethinking} and PSPNet(50)~\cite{zhao2017pyramid} to pre-train semantic segmentation models on OpenRooms, then finetune and evaluate on NYUv2~\cite{silberman2012indoor} with 40 labels~\cite{gupta2013perceptual}. We also compare the results pre-trained on InteriorNet~\cite{li2018interiornet} with the same number of training images. As shown in Tab.~\ref{tab:semantic} and Fig.~\ref{fig:segmentation}, results are comparable for the two models and register improvements with greater number of images for the two pre-training datasets.

\begin{table}
\scriptsize
\centering
\begin{tabular}{ccccccccccccc}
\cline{2-9}
\multicolumn{1}{c|}{}    & \multicolumn{4}{c|}{PSPNet(50)~\cite{zhao2017pyramid}}                                                                                                                                                                             & \multicolumn{4}{c|}{DeepLabV3~\cite{chen2017rethinking}}                                                                                                                                                                                  \\ \cline{2-9} 
\multicolumn{1}{c|}{}    & \multicolumn{2}{c|}{mIoU}                                                                      & \multicolumn{2}{c|}{mAcc}                                                                              & \multicolumn{2}{c|}{mIoU}                                                                              & \multicolumn{2}{c|}{mAcc}                                                                              \\ \cline{2-9} 
\multicolumn{1}{c|}{}    & 10K                  & \multicolumn{1}{c|}{50K}  & 10K                  & \multicolumn{1}{c|}{50K}  & 10K                  & \multicolumn{1}{c|}{50K}   & 10K                  & \multicolumn{1}{c|}{50K}  \\ \hline
\multicolumn{1}{|c|}{IN} & 41.1                  & \multicolumn{1}{c|}{41.2} & 53.3                 & \multicolumn{1}{c|}{53.4} & 41.7                 & \multicolumn{1}{c|}{42.2}  & 53.6                 & \multicolumn{1}{c|}{54.4}  \\ \hline
\multicolumn{1}{|c|}{OR} & 40.8                 & \multicolumn{1}{c|}{41.1}  & 53.0                 & \multicolumn{1}{c|}{52.5} & 42.5                 & \multicolumn{1}{c|}{42.9} & 54.5                 & \multicolumn{1}{c|}{55.1} \\ \hline
\end{tabular}
\vspace{-0.2cm}
\caption{Semantic segmentation trained on OpenRoom (OR) and InteriorNet (IN)~\cite{li2018interiornet} and fine-tuned on NYUv2~\cite{silberman2012indoor} with PSPNet(50) and DeepLabV3, using different number of images. }
\label{tab:semantic}
\vspace{-0.5cm}
\end{table}

\vspace{-0.4cm}
\paragraph{Multi-task estimation}
An advantage of OpenRooms is the ground truth available for a range of both inverse rendering and semantic tasks. This may be useful for learning  correlations among different vision tasks, and therefore can be of great interest to researchers in multi-task and transfer learning. As an illustration, we add a simple segmentation head to the inverse rendering network described above. Qualitative results are shown in Figure~\ref{fig:multitask}. Quantitative results are shown in the supplementary. We hope such data will motivate and be useful for studies in multi-task learning, such as \cite{xtc,taskonomy}.

\begin{table}[htb]
\scriptsize
\centering
\begin{tabular}{|c|c|c|c|c|}
\hline
& Barron13 \cite{barron2013intrinsic}  & Gardner17 \cite{gardner2017indoor}  & Garon19 \cite{garon2019fast} & Li20 \cite{li2020inverse} \\
\hline 
Ours vs. & 88.19\% & 66.16\% & 56.53\% & 54.77\% \\
\hline 
\end{tabular}
\vspace{-0.2cm}
\caption{User study on object insertion indicating the \% of pair-wise comparisons where human annotators thought we outperformed an alternative method; we outperform all prior methods. More details and comparisons are in supplementary.}
\label{tab:objInsertion}
\vspace{-0.5cm}
\end{table}

\vspace{-0.1cm}
\subsection{Applications to Augmented Reality}
\vspace{-0.1cm}
\paragraph{Object insertion} Photorealistic virtual object insertion in augmented reality requires high-quality estimation of geometry, material and lighting. We test our inverse network on the dataset from \cite{garon2019fast}, which contains around 80 ground-truth spatially-varying light probes. As shown in Fig.~\ref{fig:objectInsertion}, our network outperforms those methods that cannot handle spatially-varying or high-frequency lighting well. It even generates more consistent lighting compared to \cite{li2020inverse} which is trained on a SUNCG-based dataset, probably because our dataset has more diverse outdoor lighting and handles indoor lighting in a physically meaningful way. The quantitative user study in Table \ref{tab:objInsertion} also suggests that a network trained on our dataset performs better on object insertion. 

\vspace{-0.4cm}
\paragraph{Material editing} We illustrate replacement of the material of a planar surface in Fig.~\ref{fig:matReplacement} using the method of \cite{li2020inverse}. We note that spatially-varying lighting effects and specularity are handled quite well, with results comparable to  \cite{li2020inverse}, even though our dataset is created from noisy scans acquired with a commodity sensor.

\vspace{-0.1cm}
\subsection{Applications to Robotics and Embodied Vision}
\label{sec:robotics}
\vspace{-0.2cm}

To facilitate research in robotics and embodied AI, OpenRooms supports transforming a rich 3D indoor scene model into an interactive environment, with realistic physical simulation through PyBullet \cite{PyBullet}. 
A URDF file describe physical properties, such as mass and friction coefficients, for CAD models. 
This feature of OpenRooms establishes direct connections between appearance and physical properties of the environment, to provide a learning testbed for a range of topics including physics understanding from perception and policy generalization across environment and configuration changes. 
As an example, Fig.~\ref{fig:robotics} shows a classroom scenario where a robot from the iGibson dataset \cite{gibson_v1} is inserted into the scene and may perform a navigation task. Furniture in the scene can be rearranged, while the lighting and material properties can also be changed. In Fig.~\ref{fig:robotlearningtasks}, we show navigation and rearrangement where different frictions of coefficient for the same scene lead to different pushing outcomes (see supplementary for details). Since we create the scene from scans, correspondence is available to real scenes, which may be useful for sim-to-real transfer studies \cite{habitatsim2real20ral}.

\vspace{-0.4cm}
\paragraph{Ground truth for friction coefficients}
We use our albedo and roughness ground truth to render reflectance disks through a virtual equivalent of the acquisition in \cite{zhang2016friction}, then do a nearest neighbor search to compute the friction coefficients. Examples of  per-pixel friction coefficients are in Fig.~\ref{fig:friction}, where specular materials have lower friction coefficients. More details are included in the supplementary.

\vspace{-0.1cm}
\section{Conclusion and Future Work}
\label{sec:conclusion}
\vspace{-0.2cm}

We have proposed methods that enable user-generated photorealistic datasets for complex indoor scenes, starting from existing public repositories of 3D scans, shapes and materials. We illustrate the process on over 1000 indoor scenes from ScanNet. In contrast to prior works, we provide high-quality ground truth for complex materials and spatially-varying lighting, including direct and indirect illumination, light sources, per-pixel environment maps and visibility.  
We demonstrate that inverse rendering and segmentation networks can be trained on OpenRooms, towards augmented reality applications like object insertion and material editing. We also show our dataset can be integrated with physics engines and provide friction coefficients, which suggest interesting future studies in navigation, rearrangement and sim-to-real transfer.
Our dataset and all tools used for its creation will be publicly released.

Please refer to the {\bf supplementary material} for further details, extensive experimental results and videos.

\vspace{0.1cm}

\small
\noindent\textbf{Acknowledgments: }
We thank NSF CAREER 1751365, a Google Award, generous gifts from Adobe, NSF CHASE-CI, ONR N000142012529, N000141912293, NSF 1703957, the Ronald L. Graham Chair and UCSD Center for Visual Computing. We thank Fei Xia for helpful suggestions.
\normalsize

\appendix 
\section{Summary of the Supplementary Material}
Our supplementary material includes an accompanying video, further details of the dataset and tools, as well as further results and comparisons on several tasks and applications.
This supplementary document includes the following:
\begin{tight_itemize}
\item Explanation of the accompanying video (Sec.~\ref{sec:video})
\item Demonstrations of editing applications (Sec.~\ref{sec:editing})
\item Further results on inverse rendering (Sec.~\ref{sec:inverse})
\item Ground truth for friction coefficients (Sec.~\ref{sec:friction})
\item Demonstrations to motivate robotic tasks (Sec.~\ref{sec:robotics_sup})
\item Results on semantic and instance segmentation (Sec.~\ref{sec:segmentation})
\item Multi-task estimation and domain adaptation (Sec.~\ref{sec:segmentation})
\item Photorealistic ground truth for SUN-RGBD (Sec.~\ref{sec:SUNRGBD})
\item Details of the microfacet BRDF model (Sec.~\ref{sec:BRDF})
\item Details of the lighting ground truth (Sec.~\ref{sec:lighting})
\item Details of physically-based GPU renderer (Sec.~\ref{sec:renderer}).
\end{tight_itemize}

\section{Video}
\label{sec:video}

The accompanying video is included at the following \href{https://ucsd-openrooms.github.io/
}{link}. It illustrates the following capabilities enabled by the proposed OpenRooms framework:
\begin{tight_itemize}
\item Creating photorealistic synthetic versions of real acquired scans, with side-by-side comparisons
\item Beyond shape, extensive ground truth for high-quality spatially-varying material and spatially-varying lighting with various elements of complex light transport
\item Ground truth for semantic and instance segmentation
\item Ground truth for friction coefficients
\item Inverse rendering and scene understanding applications
\item Image editing applications for augmented reality
\item Motivation for robotics applications such as navigation, pushing and sim-to-real transfer studies, where variations in material and lighting may be important.
\end{tight_itemize}
The video also illustrates various steps of the proposed dataset creation framework, where besides the images and ground truth, the involved tools are being publicly released as part of our open framework.

\section{Applications: Photorealistic Image Editing}
\label{sec:editing}

\begin{table}[!!t]
\scriptsize
\centering
\begin{tabular}{|c|c|c|c|c|}
\hline
    Barron13 \cite{barron2013intrinsic}  & Gardner17 \cite{gardner2017indoor}  & Garon19 \cite{garon2019fast} & ~~Li20 \cite{li2020inverse}~~  & ~~~Ours~~~ \\
\hline
11.5\% & 28.07\% & 29.15\% & 34.84\% & \color{blue}38.89\%\\
\hline 
\end{tabular}
\caption{User study on object insertion by comparing to the ground-truth. Here we compare the lighting prediction results of different methods against ground truth lighting and report the \% of times that users picked a particular method as being more realistic than ground truth; ideal performance is 50\%. Our result is marked in ({\color{blue}{blue}}). Similar to the results in the main paper, our trained network outperforms previous state-of-the-art ones.}
\vspace{-0.4cm}
\label{tab:objInsertion_vsGT}
\end{table}

\begin{table}[!!t]
\scriptsize
\centering
\begin{tabular}{|c|c|c|c|c|c|}
\hline
   & \cite{barron2013intrinsic}  & \cite{gardner2017indoor}  &  \cite{garon2019fast} & \cite{li2020inverse}  & Ours \\
\hline
 Barron13 \cite{barron2013intrinsic} & - & 23.37\% & 13.25\% & 13.60\% & \color{blue}11.81\% \\
 \hline 
 Gardner17 \cite{gardner2017indoor} & 76.63\% & - & 36.25\% & 39.54\% & \color{blue}33.84\%  \\
 \hline
 Garon19 \cite{garon2019fast} & 86.75\% &  63.75\% & - & 42.28\% & \color{blue}43.47\%   \\
 \hline 
 Li20 \cite{li2020inverse} & 86.40\% & 60.46\% & 57.72\% & - & \color{blue}45.23\%   \\
 \hline 
 Ours & \color{blue}88.19\% & \color{blue}66.16\% & \color{blue}56.53\% & \color{blue}54.77\% & - \\
 \hline 
 \end{tabular}
 \caption{User study on object insertion with pairwise comparisons. $X$\% in row $I$ column $J$ means that in $X$\% of total cases, human annotators think method $I$ outperforms method $J$. Comparisons with our method are labeled in ({\color{blue}{blue}}). We observe improvements over all prior methods.}
 \label{tab:objInsertion_sup}
 \vspace{-0.4cm}
 \end{table}

\begin{figure*}
\centering
\includegraphics[width=\textwidth]{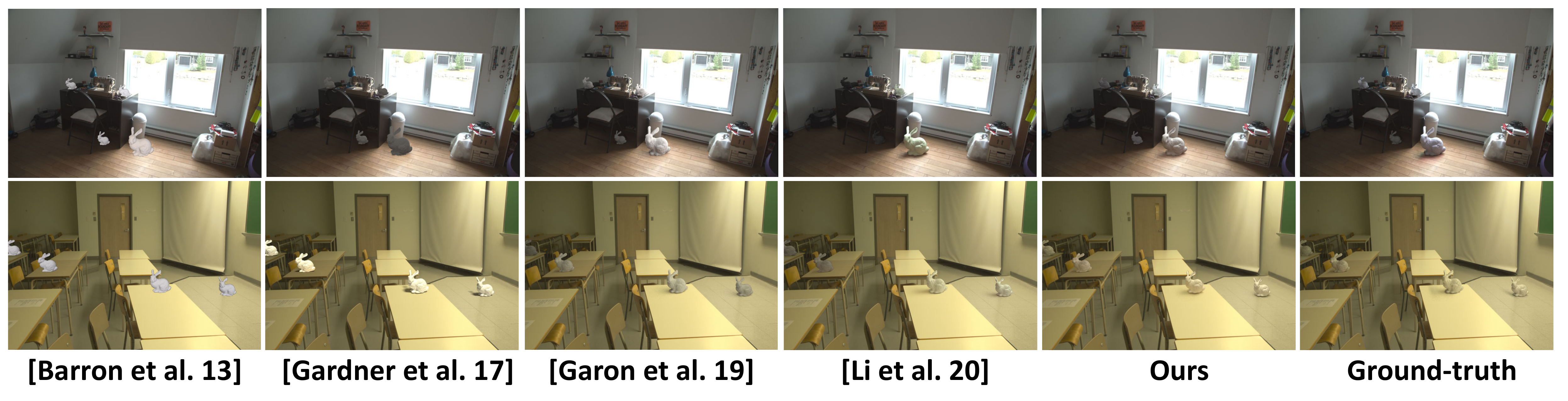}
\caption{Qualitative comparisons of object insertion on real images from the dataset of \cite{garon2019fast}. }
\label{fig:objectInsertion_supp}
\end{figure*}

\paragraph{User study: Object insertion} A user study was conducted to quantitatively evaluate object insertion performance using the inverse rendering network of the main paper. The network is trained on the proposed dataset and evaluated on the real dataset of \cite{garon2019fast}, which provides 20 images with measured ground truth for spatially-varying lighting. Some qualitative and quantitative results have been included in Table 6 and Figure 13 of the main paper. We now provide more comparisons in Table \ref{tab:objInsertion_vsGT}, Table \ref{tab:objInsertion_sup} and Figure \ref{fig:objectInsertion_supp}. 

In Table \ref{tab:objInsertion_vsGT}, we summarize comparisons for different methods against ground-truth lighting. Ideal performance for this task is 50\%, which indicates that the predicted lighting and the ground-truth lighting are indistinguishable. The best two previous methods of \cite{li2020inverse} (34.84\%) and \cite{garon2019fast} (29.15\%) are trained on SUNCG-related datasets, while our method (38.89\%) outperforms both of them. In Table \ref{tab:objInsertion_sup}, we show complete pairwise comparisons for object insertion among recent state-of-the-art lighting prediction methods. This is a more detailed version of Table 6 in the main paper and reaffirms that a network trained on the proposed dataset achieves the best performances. In Figure \ref{fig:objectInsertion_supp}, we show more qualitative comparisons. The network trained on our dataset achieves realistic high frequency shading and consistent lighting color. 

In conclusion, the dataset created by our framework enables high-quality object insertion with performance better than methods built on previous datasets.

\begin{figure}
\centering
\includegraphics[width=\columnwidth]{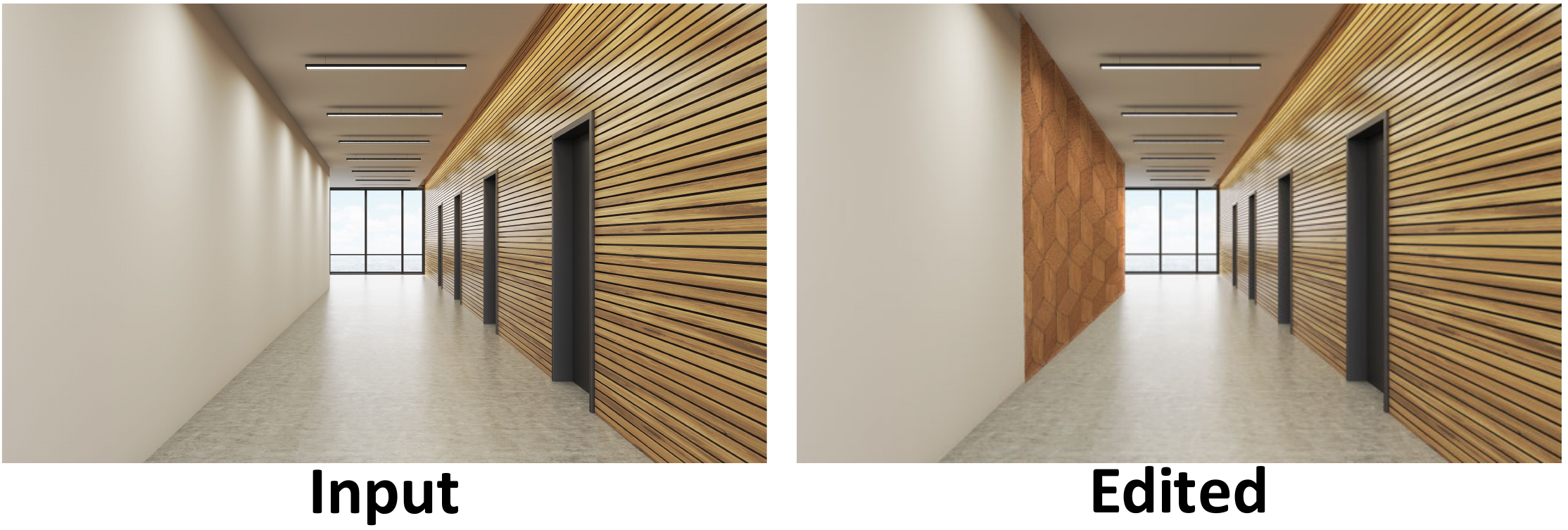}
\vspace{-0.3cm}
\caption{Material replacement on a real image using the inverse rendering predictions from a network trained on our dataset.}
\label{fig:matEdit}
\vspace{-0.4cm}
\end{figure}

\vspace{-0.4cm}
\paragraph{\bf Qualitative results: material editing}
We show material replacement examples on real images in Figure 14 in the main paper and Figure \ref{fig:matEdit} here. Since we use a per-pixel environment map to represent spatially-varying lighting, we can recover complex spatially-varying highlights when we replace the original material with another glossy material. 

\section{Inverse Rendering Trained on OpenRooms}
\label{sec:inverse}

This section includes: {\bf (a)} further results on light source detection, {\bf (b)} quantitative results on per-pixel lighting estimation, {\bf (c)} comparisons of normal estimation with prior works on real datasets, {\bf (d)} comparisons for layout estimation, {\bf (e)} ablation study for the network on our proposed dataset, {\bf (f)} qualitative visualization of inverse rendering network outputs on synthetic and real data, when trained on a synthetic dataset created from ScanNet using the proposed dataset creation method.

\begin{table}[t]
\centering
\setlength{\tabcolsep}{4pt}
\scriptsize
\begin{tabular}{|c|c|c|c|c|c|}
\hline
 & $A$($10^{-3}$) & $N$($10^{-2}$) & $D$($10^{-2}$) & $R$($10^{-2}$) & $L$  \\
 \hline 
Cascade0 & 9.99  & 4.51 & 5.18 & 6.59 & 0.150 \\
\hline 
Cascade1 & 9.43 & 4.42 & 4.89 & 6.64 & \textbf{0.146} \\
\hline 
Bilateral solver & \textbf{9.29} & - & \textbf{4.86} & \textbf{6.57} & - \\
\hline
\end{tabular}
\vspace{-0.3cm}
\caption{Ablation study for the network architecture on our proposed dataset. We report the scale invariant L2 loss for albedo ($A$), L2 loss for normal ($N$), scale invariant $\log$ L2 loss for depth ($D$), L2 loss for roughness ($R$) and scale invariant $\log(x+1)$ L2 loss for per-pixel lighting ($L$). We observe both cascade structure and bilateral solver can improve the prediction accuracy.  }
\label{tab:ablation}
\end{table}

\begin{figure}
\centering
\includegraphics[width=\linewidth]{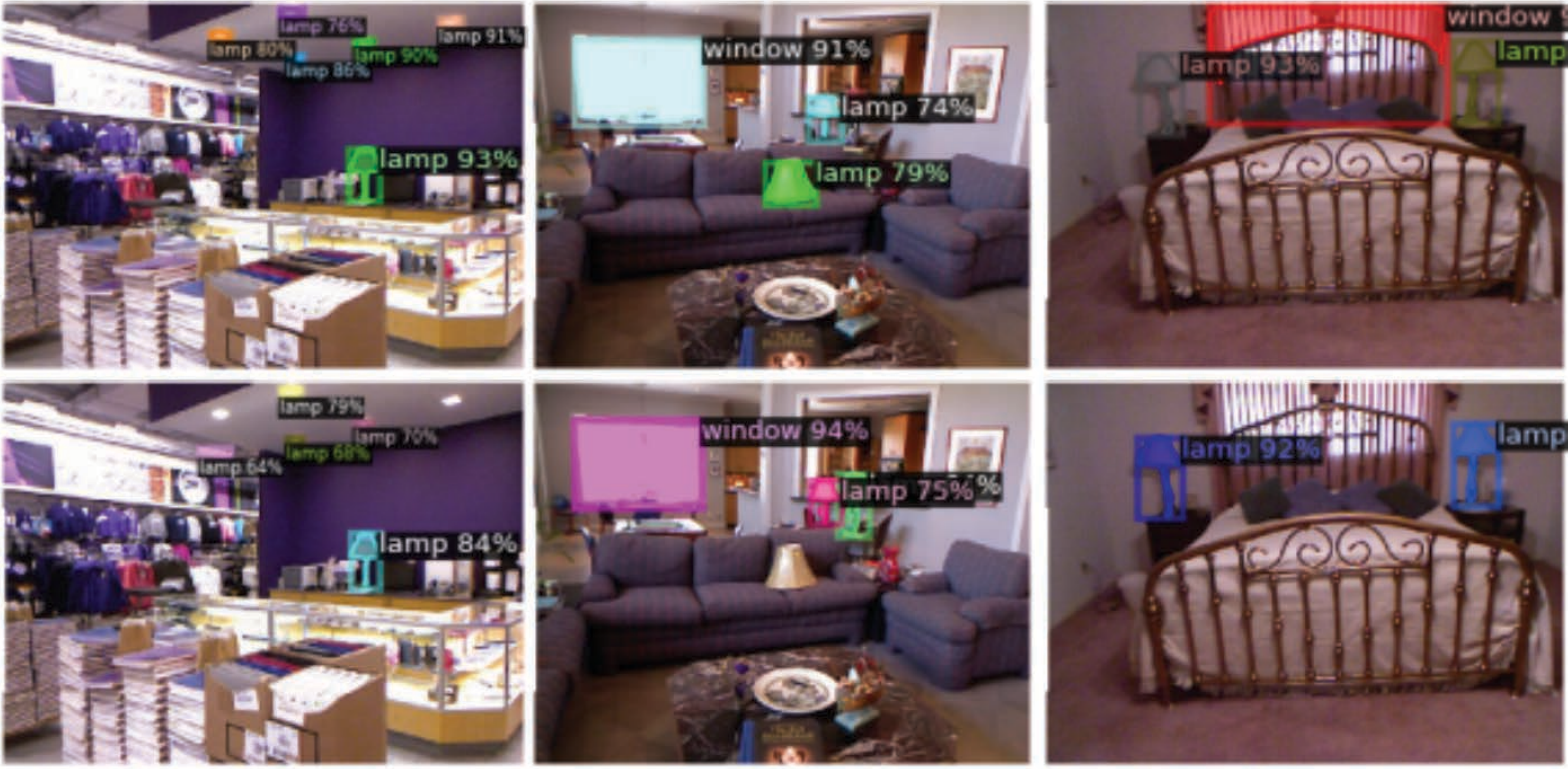}
\vspace{-6mm}
\caption{Further results of light source detection on NYUv2 test images. The top row is with pre-training on OpenRooms and the bottom row without the pre-training.}
\label{fig:light_detect_sup}
\vspace{-0.4cm}
\end{figure}

\begin{figure*}[t]
\centering
\includegraphics[width = \textwidth]{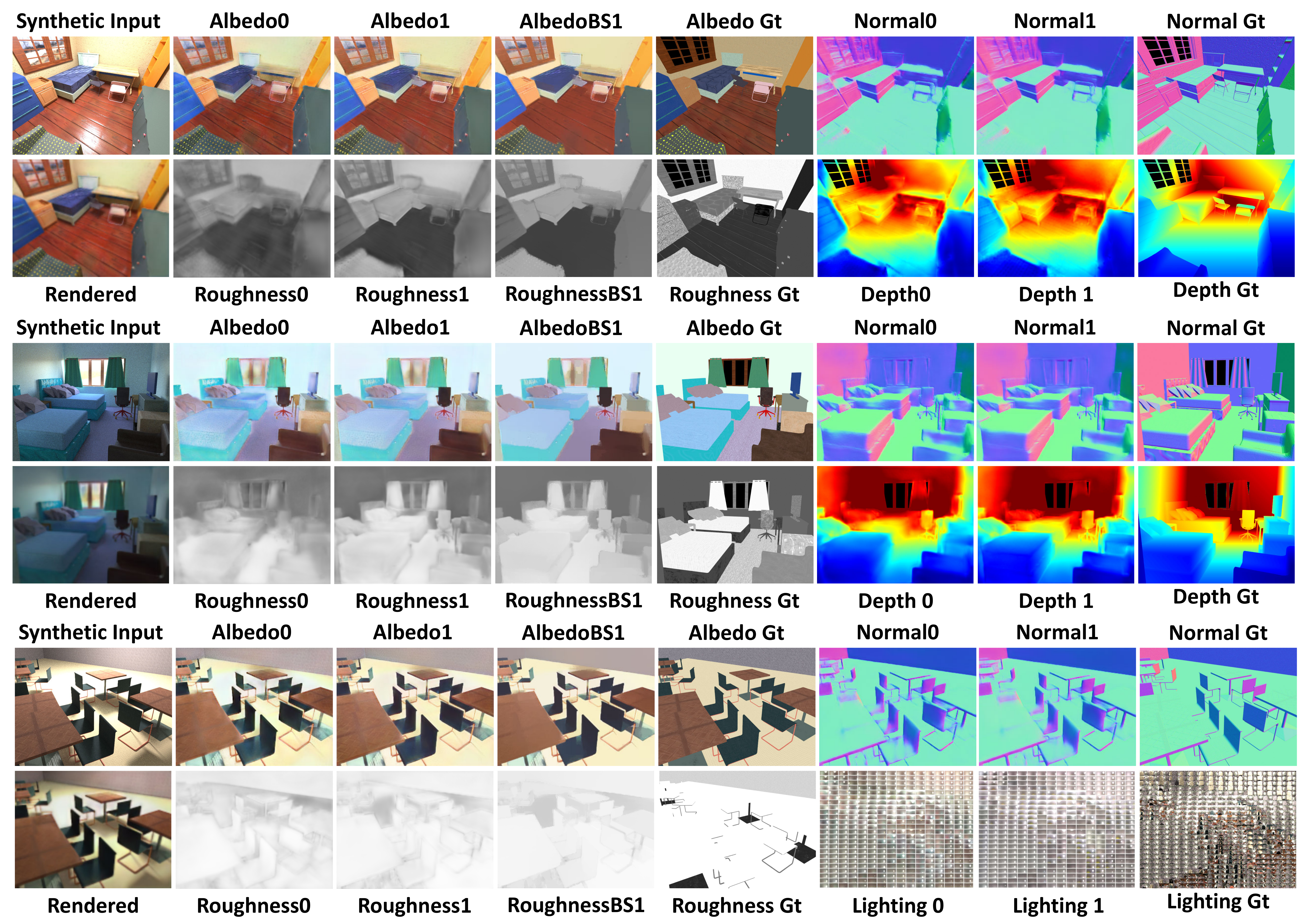}
\caption{Qualitative visualization of inverse rendering results on synthetic images from the test set of the proposed dataset. }
\label{fig:synthetic}
\end{figure*}

\vspace{-0.4cm}
\paragraph{\bf Inverse rendering on test set of proposed dataset} 
Table \ref{tab:ablation} quantitatively evaluates the performance of the network trained and then tested on the proposed synthetic dataset created from ScanNet. We observe that both the cascade structure and bilateral solver can improve the accuracy of prediction of most intrinsic components. Figure \ref{fig:synthetic} shows a few inverse rendering results on our synthetic testing set. From the figure, we observe that through iterative refinement, the cascade structure can effectively remove noise and recover high-frequency signals, especially for lighting and normal prediction. The bilateral solver also helps remove noise by enhancing the smoothness prior.

\begin{figure*}[t]
\centering
\includegraphics[width=\textwidth]{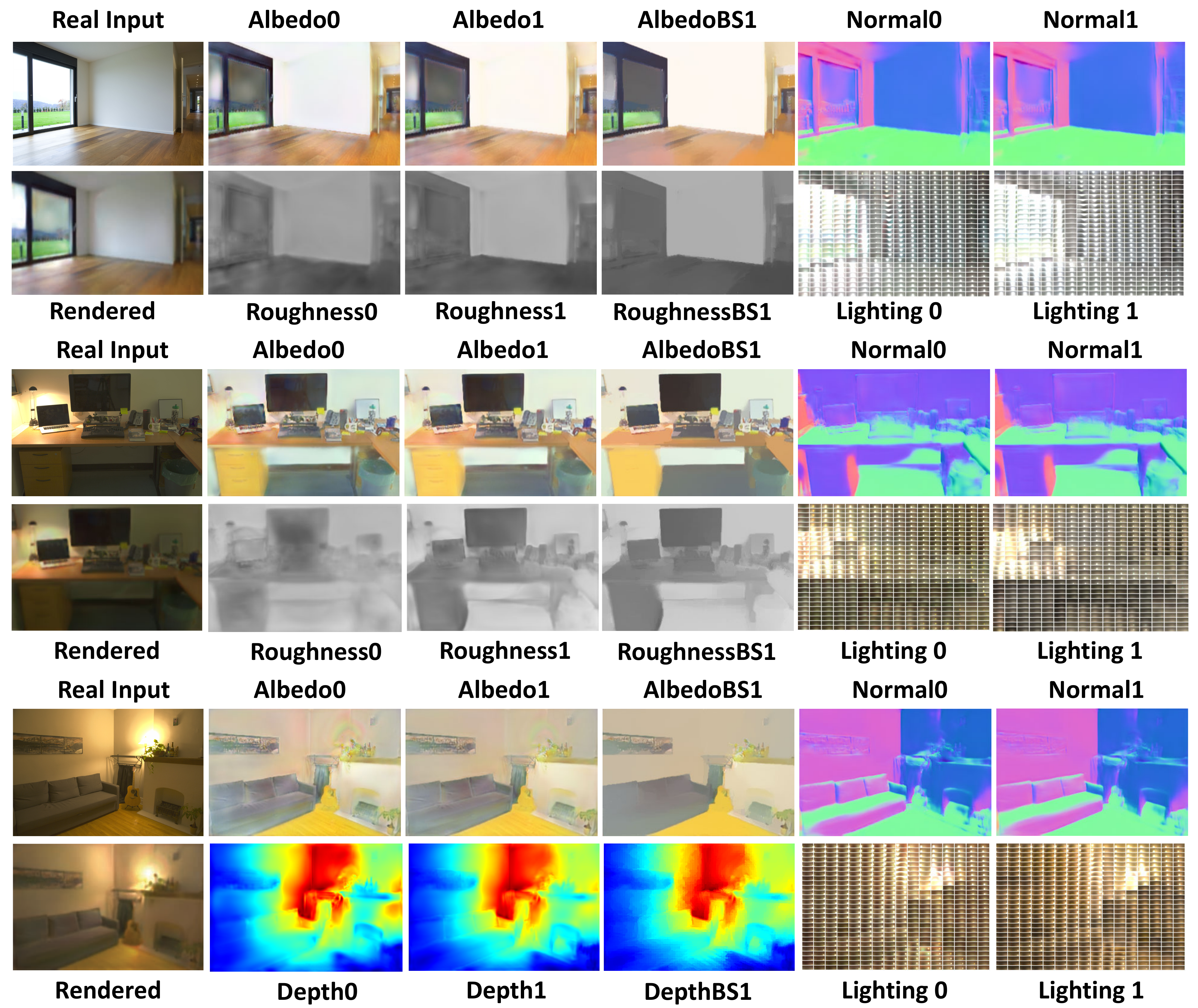}
\caption{Qualitative visualization of inverse rendering results on real images, using a network trained on synthetic photorealistic images from the proposed OpenRooms dataset (created based on scans from ScanNet).}
\label{fig:real_supp}
\end{figure*}

\paragraph{\bf Further examples of inverse rendering on real images} Figure \ref{fig:real_supp} shows inverse rendering results on several real images. We observe that even though the network is trained on a synthetic dataset, it can generalize well to real data. For real data, the effectiveness of the cascade structure and bilateral solver is more apparent, probably due to noisier initial predictions on real data.

\vspace{-0.4cm}
\paragraph{Light source detection}
We include further examples for light source detection in Figure \ref{fig:light_detect_sup}, besides the quantitative numbers in Table 4 and visualizations in Figure 10 of the main paper. We again observe that pre-training on OpenRooms is beneficial for detecting both wondows and lamps.

\vspace{-0.4cm}
\paragraph{Per-pixel lighting estimation}
We have shown per-pixel lighting prediction results on both real and synthetic examples in Figure 9 of the main paper. We now provide further qualitative results on real and synthetic data in Figure \ref{fig:real_supp} and Figure \ref{fig:synthetic}, respectively. Table \ref{tab:ablation} shows quantitative numbers on our OpenRooms validation set. We observe that our per-pixel lighting prediction is consistent with spatially-varying intensity and the ground-truth light source position. Both quantitative and qualitative comparisons show that cascade structure can improve the per-pixel lighting prediction by making the prediction sharper and less noisy.

\begin{figure*}[t]
\centering
\includegraphics[width=0.9\textwidth]{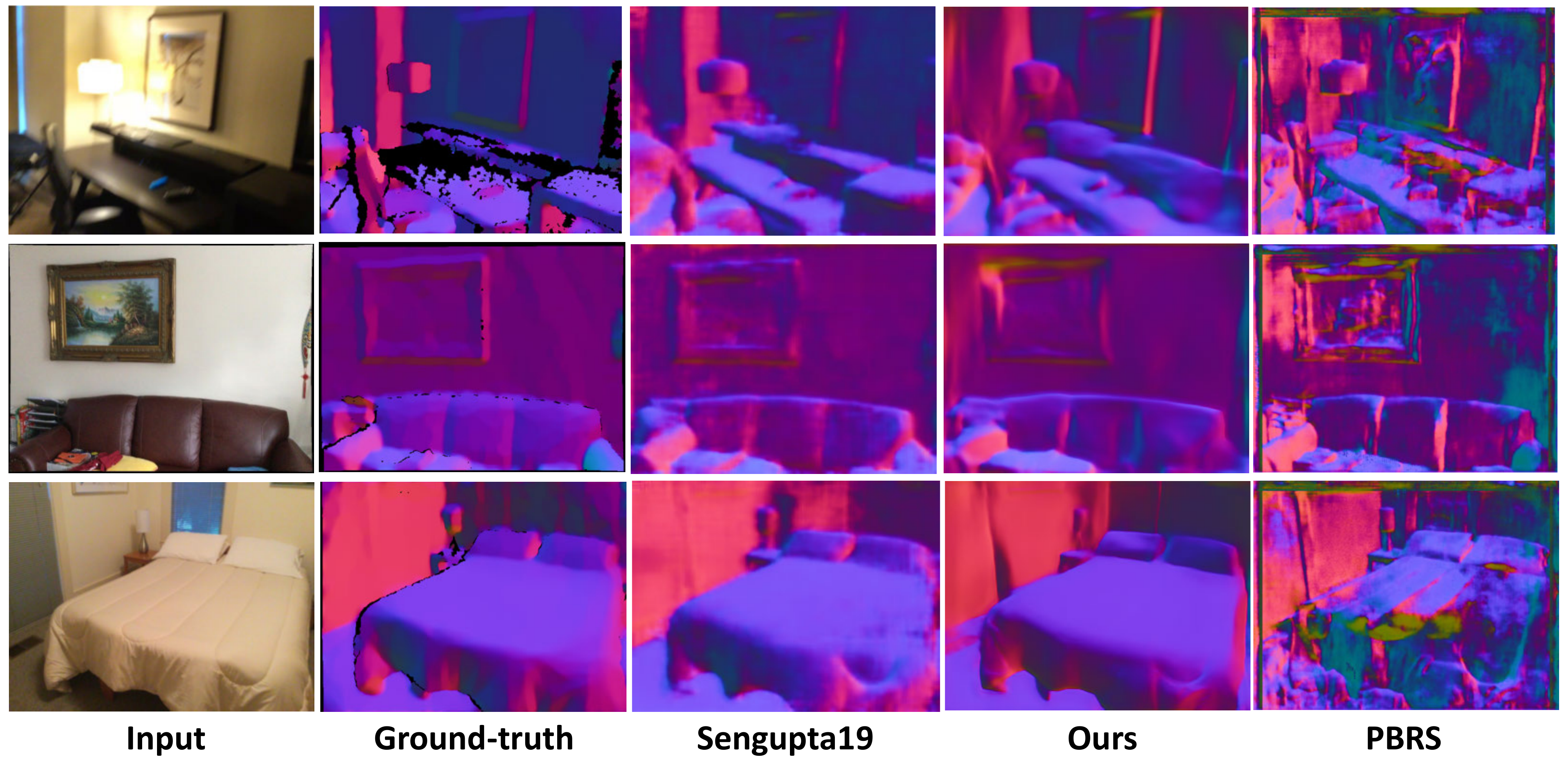}
\caption{Qualitative comparisons of our normal estimation with Sengputa19 \cite{sengupta2019neural} and PBRS \cite{zhang2016physically}, on real images from \cite{sengupta2019neural}. }
\label{fig:normal}
\end{figure*}

\vspace{-0.4cm}
\paragraph{\bf Normal prediction} 
Figure \ref{fig:normal} shows qualitative comparisons with \cite{sengupta2019neural} and \cite{zhang2016physically} on three real examples from  \cite{sengupta2019neural}. We observe that even though \cite{zhang2016physically} achieves the best accuracy on NYU dataset, it might overfit to that specific dataset and might not generalize well to images from other sources. On the contrary, both \cite{sengupta2019neural} and our network achieve less noisy normal predictions. Our network may sometimes over-smooth the normal, probably since our scenes are built from Scan2CAD annotations that usually contain only a small number of large items of furniture in each room. Therefore, there may be less geometric detail in our synthetic dataset. This can probably be solved in the future by procedurally adding small objects to the rooms to increase the complexity of the dataset.

\begin{table}[t]
\centering
\scriptsize
\begin{tabular}{|c|c|c|c|c|c|}
\hline
    \multicolumn{1}{|c}{} & \multicolumn{2}{|c|}{Corner} & \multicolumn{2}{c|}{Edge} & \multicolumn{1}{c|}{Room} \\
    \hline
     & Precision & Recall & Precision & Recall & IOU \\ 
     \hline
     Chen19 \cite{chen2019floor} & 0.358 & 0.524 & 0.151 & 0.191 & 0.734 \\
     Trained on ScanNet & 0.531 & 0.716 & 0.254 & 0.316 & 0.858 \\ 
\hline 
\end{tabular}
\caption{Comparison of Floor-SP \cite{chen2019floor} models with pre-trained weights provided by \cite{chen2019floor} and weights trained on 1069 ScanNet scenes. The network trained on our newly labeled scenes perform significantly better on noisy scanned point cloud.}
\label{tab:layoutPrediction}
\end{table}

\begin{figure*}[t]
\centering
\includegraphics[width=0.9\textwidth]{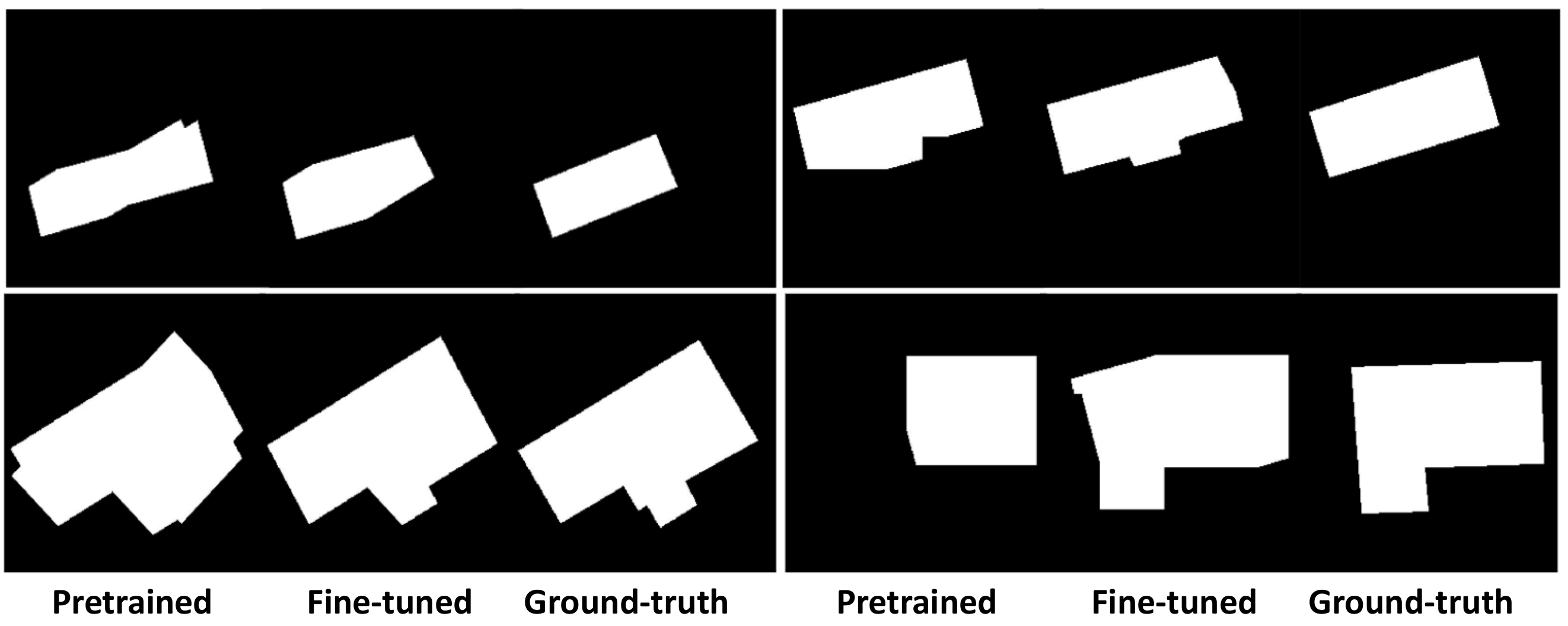}
\caption{Comparison of layout reconstruction using the original network from \cite{chen2019floor} and the network trained on our ScanNet annotation. }
\label{fig:layout}
\end{figure*}

\vspace{-0.4cm}
\paragraph{\bf Layout prediction} 
For future ease of annotation, we add an automatic layout predictor using Floor-SP \cite{chen2019floor} to the OpenRooms tools. It accepts a 2D top-down projection of the point cloud and its mean surface normal as inputs. In the subsequent steps, the room segmentation is predicted and room loops are formed (we omit the loop merging step since ScanNet scans generally contain a single room). We refer the reader to \cite{chen2019floor} for more details. Since the point cloud generated by RGB-D scans contain higher levels of noise compared to the training data used by Floor-SP, we trained a randomly initialized model on a subset of ScanNet consisting of 1069 scenes with human-annotated layout as ground-truth. 

The final layout is evaluated on 103 held-out scenes in terms of corner precision and recall, edge precision and recall, as well as intersection-over-union of the room segmentation. A corner prediction is deemed correct if its distance to the closest ground truth corner is within 10 pixels. An edge prediction is deemed valid if its two endpoints pass the criterion for corners and the edge belongs to the set of ground-truth edges. 

Table \ref{tab:layoutPrediction} shows the comparison between the model trained on ScanNet and the pre-trained weights provided by the original implementation of Floor-SP. Figure \ref{fig:layout} shows a qualitative comparison. Note that the room segmentation performs moderately well despite the low precision and recall of the corners and edges. We believe that this is caused by the ambiguities during layout annotation. Since we require the walls to be arranged such that they form a closed loop, for the scans that do not cover the entire room, the human annotator would have to add false corners and edges that pass through open areas where the scan is incomplete, thereby affecting the evaluation of the corner and edge predictions. On the other hand, false corners and edges do not affect IoU since it measures the area covered by the room, rather than the occurrence of predictions.

\section{Ground Truth for Friction Coefficients}
\label{sec:friction}

In this section, we describe in detail the process for assigning per-pixel ground truth for friction coefficients in OpenRooms scenes mentioned in Sec 4.3 of the main paper. We also show several qualitative examples for assigned friction coefficients. Since OpenRooms provides complete control over mapping arbitrary semantically-meaningful materials to indoor surfaces, such ground truth may enable future studies in inferring physical properties from images, or learning robotics tasks conditioned on material properties.

\begin{figure}
\centering
\includegraphics[width=\columnwidth]{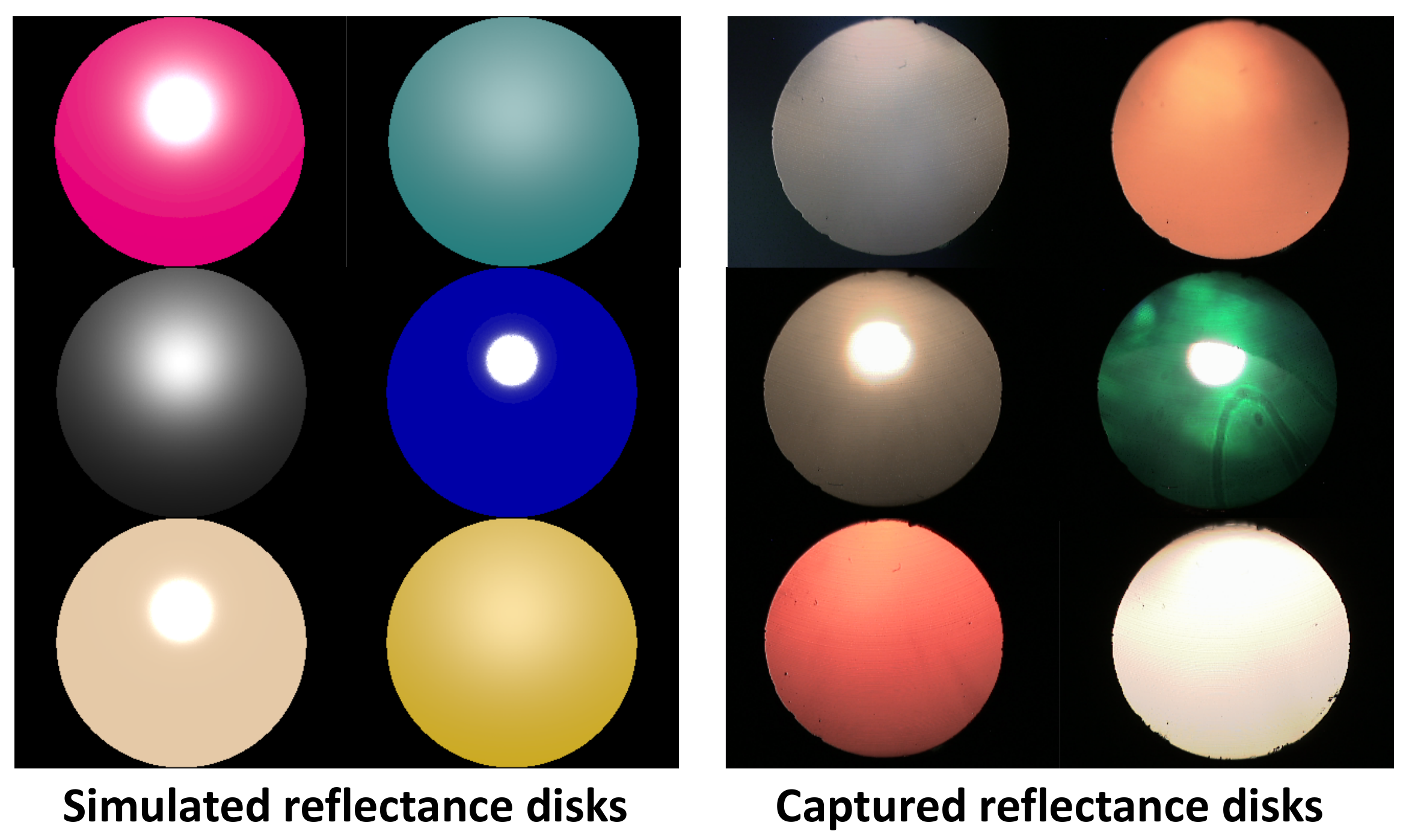}
\caption{Comparisons of randomly sampled reflectance disks captured by the system of Zhang et al.~\cite{zhang2016friction} (left) and rendered by our virtual environment (right). We observe the distributions of highlights and spatially-varying intensities to be similar.}
    \label{fig:rfDisk}
\end{figure}

\begin{figure}
\centering
\includegraphics[width=\columnwidth]{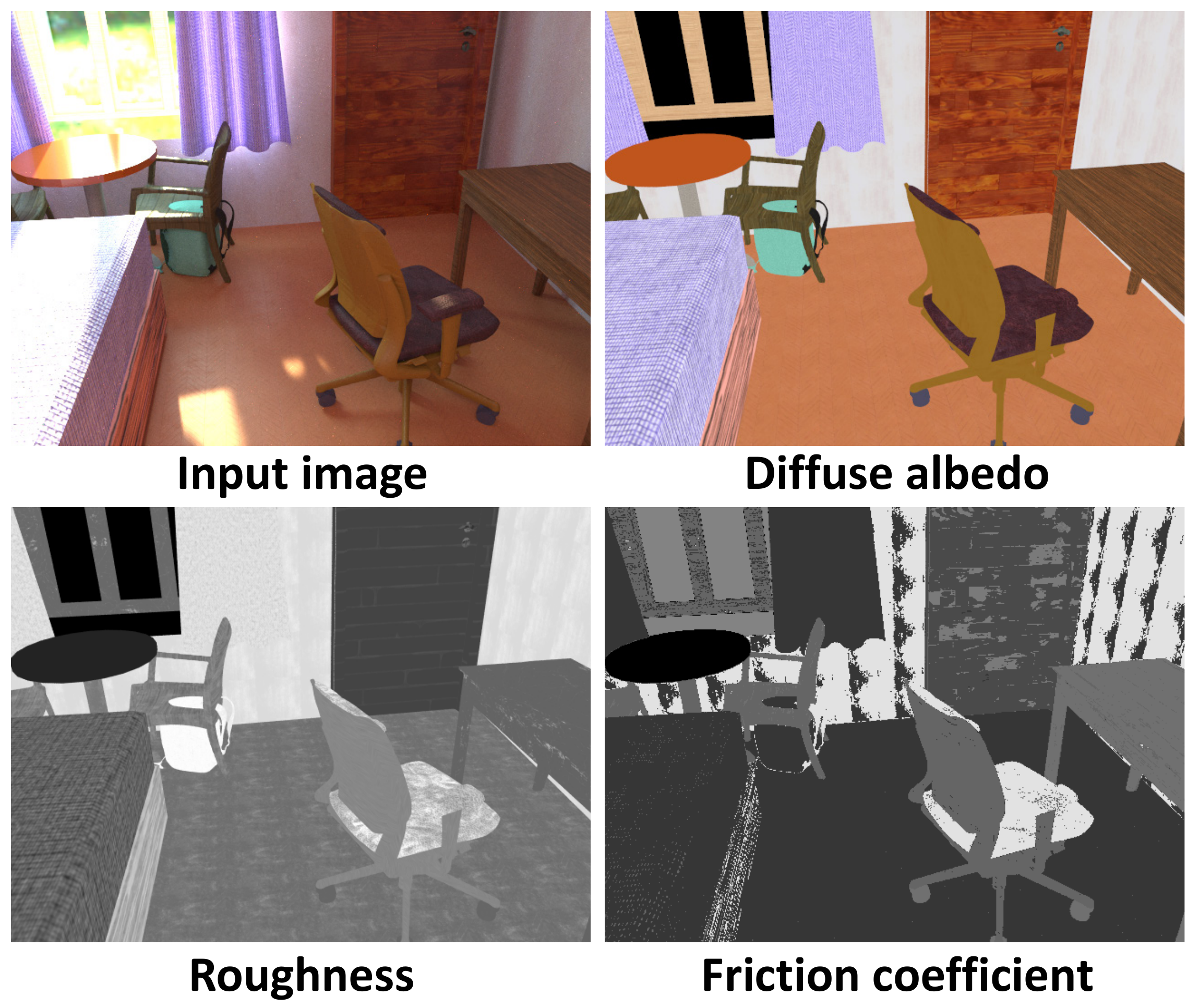}
\caption{Visualization of friction coefficient in OpenRooms dataset. We map diffuse albedo and roughness parameters to friction coefficient based on nearest neighbor search. We observe that specular materials usually have smaller friction coefficients.  }
\label{fig:friction_sup}
\end{figure}

\vspace{-0.4cm}
\paragraph{Reflectance disks} 
We follow the concept of reflectance disks from Zhang et al.~\cite{zhang2016friction} for predicting friction coefficients for various materials. 
The acquisition setup of \cite{zhang2016friction} includes a beam splitter, an orthographic camera and a parabolic mirror, to capture material appearances by densely sampling from a large range of view directions and a small range of lighting directions (please see Figure 3 of \cite{zhang2016friction}). We mimic this capture system to render the reflectance disk using our physically-based renderer. We uniformly sample the parameter space of our microfacet BRDF model and render a reflectance disk for each sampled point. Figure \ref{fig:rfDisk} compares the reflectance disks rendered under our virtual environment and captured by the system. We observe that the distribution of specular highlights and intensities of the two sets of reflectance disks can match well. 

\vspace{-0.4cm}
\paragraph{Deep reflectance codes} After obtaining the reflectance disk, Zhang et al.~\cite{zhang2016friction} use a pretrained deep network to map the reflectance disk to a low dimensional latent space, which is termed a deep reflectance code. Due to the dense down-sampling operations, the deep reflectance code is robust to translation and rotation, which makes it a suitable representation for modeling intrinsic properties of materials, including the friction properties. Thereafter, they use K-nearest neighbor method to map deep reflectance code to friction coefficients. Following their implementation, we also map our reflectance disks to a deep reflectance code, to the friction coefficients using nearest neighbor search for each of our sampled microfacet BRDF parameters. This gives us a table that allows us to map our microfacet BRDF parameters to friction coefficients through bilinear interpolation or nearest neighbor search. Figure 16 in the main paper and Figure \ref{fig:friction_sup} in the supplementary show some examples of our friction coefficient predictions. We observe that specular materials are more likely to have small coefficients of friction, which is consistent with physical intuition.

\makeatletter
\def\thickhline{%
  \noalign{\ifnum0=`}\fi\hrule \@height \thickarrayrulewidth \futurelet
   \reserved@a\@xthickhline}
\def\@xthickhline{\ifx\reserved@a\thickhline
               \vskip\doublerulesep
               \vskip-\thickarrayrulewidth
             \fi
      \ifnum0=`{\fi}}
\makeatother

\newlength{\thickarrayrulewidth}
\setlength{\thickarrayrulewidth}{2\arrayrulewidth}

\begin{figure*}[t]
\centering
\includegraphics[width = 0.9\textwidth]{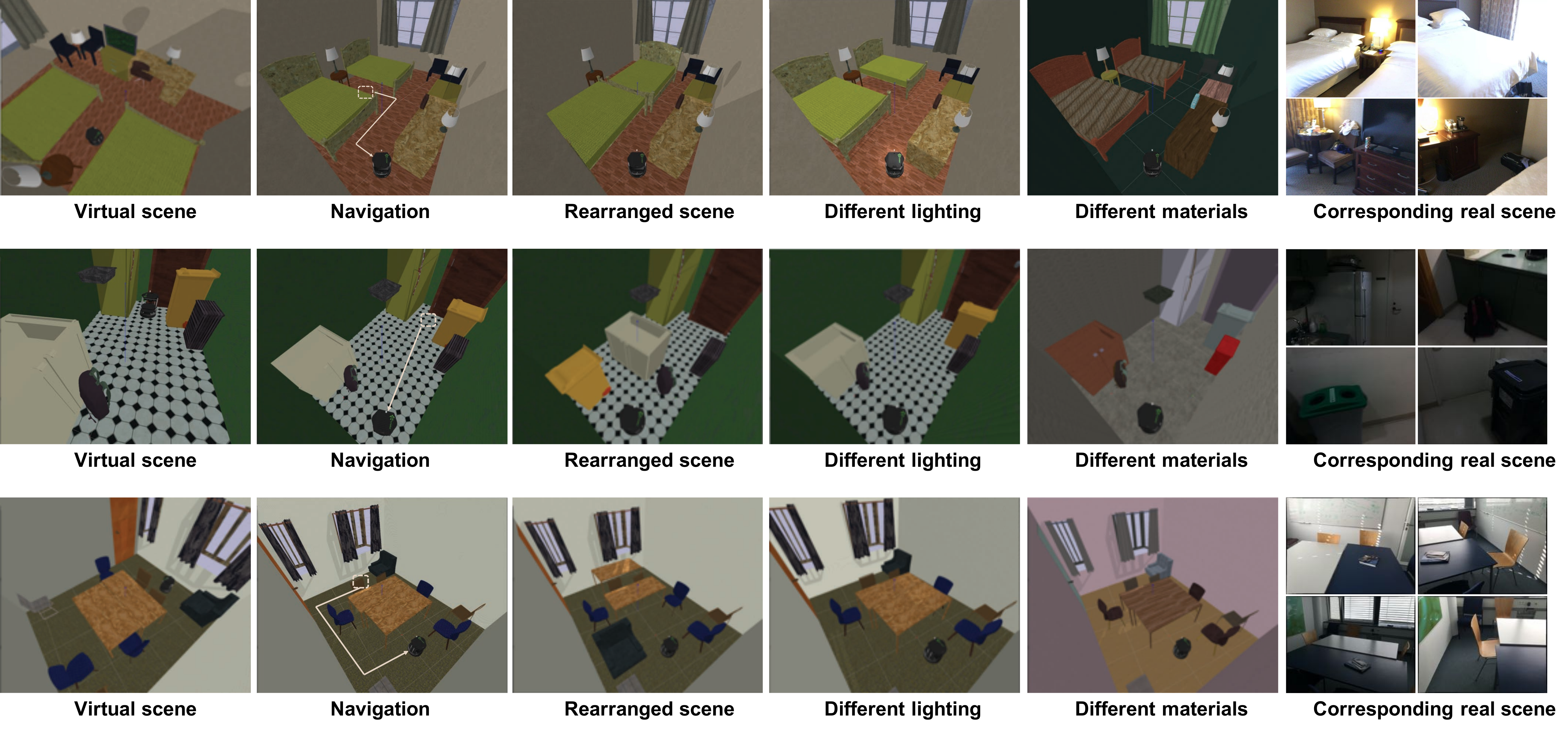}
\vspace{-0.3cm}
\caption{More examples (bedroom, kitchen, and conference room) of OpenRooms scenarios integrated with a physics engine under different settings, as well as the images from the corresponding real scenes.}
\label{fig:robotics_qualitative_examples}
\vspace{-0.3cm}
\end{figure*}

\begin{figure}[t]
\centering
\includegraphics[width=\linewidth]{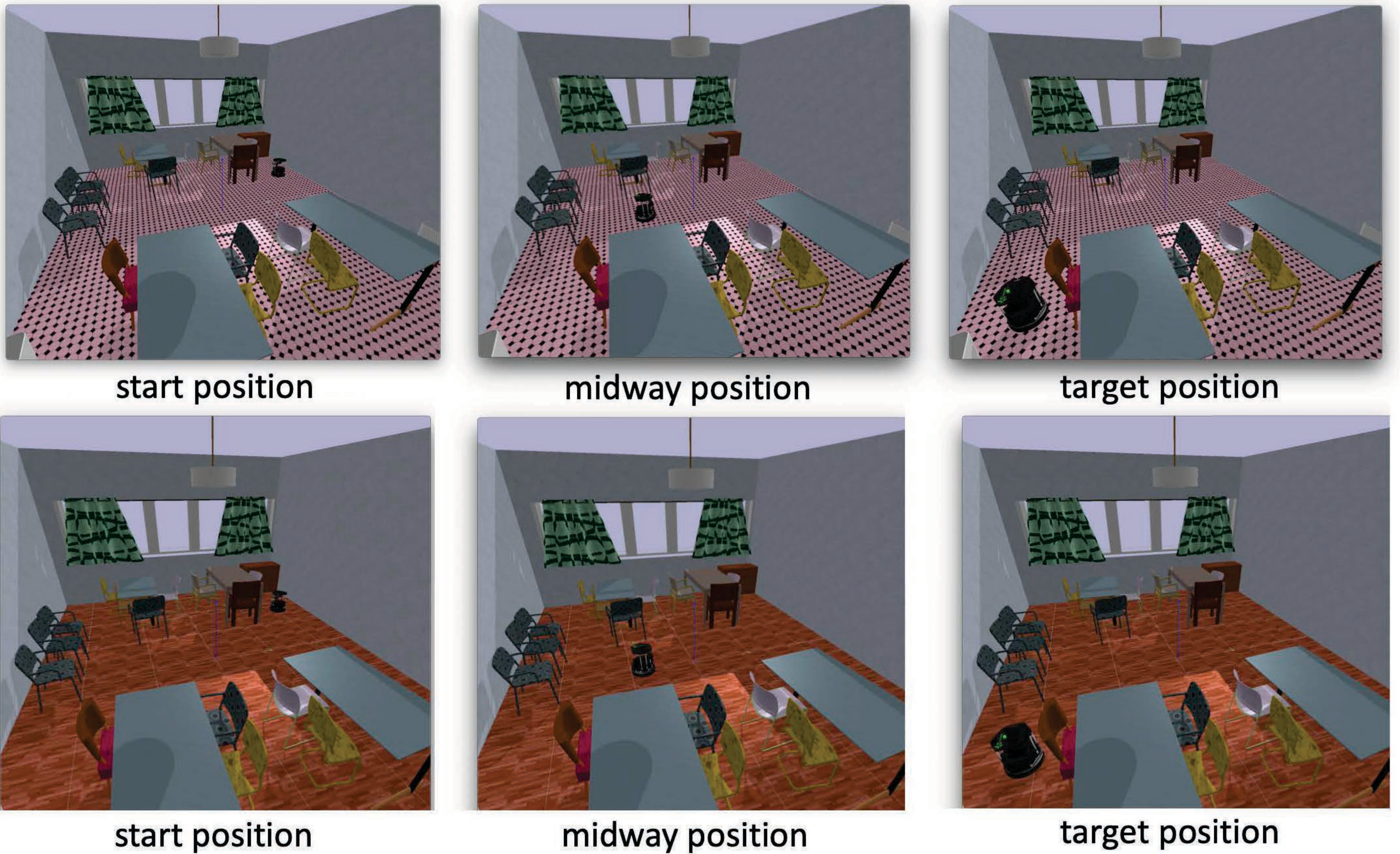}
\vspace{-0.3cm}
\caption{A Turtlebot navigating in a classroom from the brown cabinet to the pink chair, on floors with different materials.}
\label{fig:navigation}
\vspace{-0.3cm}
\end{figure}

\begin{figure}[t]
\centering
\includegraphics[width=\linewidth]{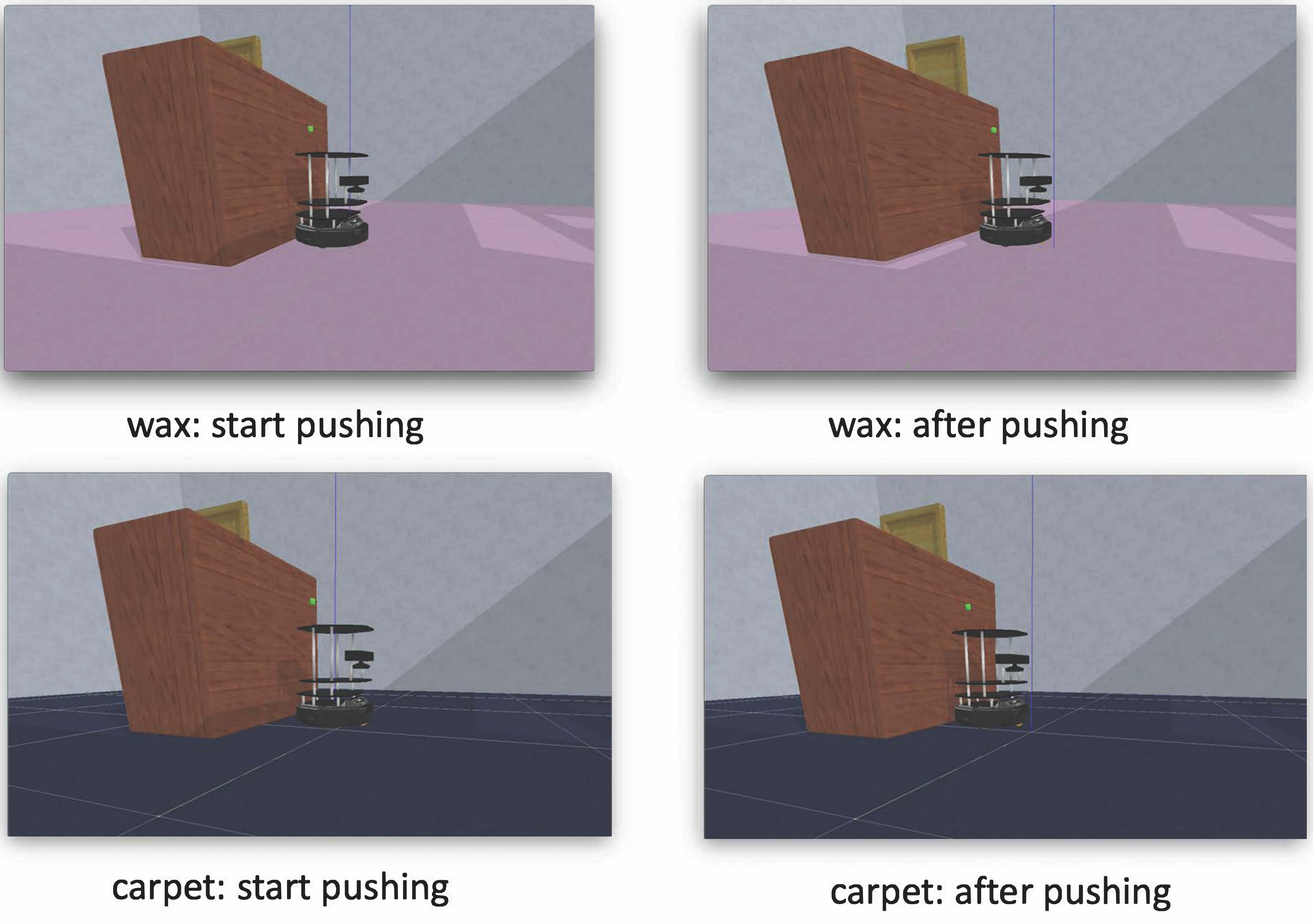}
\vspace{-0.3cm}
\caption{A Turtlebot pushing a wooden cabinet on floors with materials of different friction coefficients: wax and carpet, leading to differing physical outcomes (also see the accompanying video).}
\label{fig:pushing}
\vspace{-0.3cm}
\end{figure}

\section{Applications: Robotics, Embodied Vision}
\label{sec:robotics_sup}

In this section, we provide details and examples for the following: {\bf (a)} integration of OpenRooms scenes with PyBullet for physical simulation, {\bf (b)} qualitative results of OpenRooms scenes and capabilities enabled by such integration, {\bf (c)} demonstration of navigation in OpenRooms scenes, {\bf (d)} demonstration of pushing tasks with different coefficients of friction.

\vspace{-0.4cm}
\paragraph{Integrating OpenRooms with PyBullet}
To transform a static OpenRooms scene to an interactive environment, we treat each object in the scene as a single link robot and equip it with a URDF to describe its physical and visual properties. In our dataset, the object's 3D mesh and the associated MTL file are recorded in an OBJ file. Given this OBJ file, we generate another OBJ file by convex decomposition. The URDF links these two OBJ files, using the first one for rendering and the other for collision detection. From the albedo and roughness images provided in the MTL file, we can estimate the object's friction coefficient. Other physical properties, such as the mass, center of mass and inertial matrix can also be provided in the URDF or set in the physics engine later. Having URDFs for each object in the scene, we then load them along with the robot's URDF into the physics engine (for example, PyBullet) to allow full interactive physics simulations.

\vspace{-0.4cm}
\paragraph{Qualitative examples}
All kinds of OpenRooms scenarios can be integrated with the physics engine (Pybullet in our case) to create interactive environments where a robot can act (for example, navigate or push objects). The objects can be rearranged, the light sources replaced and the materials changed, boosting the variety of the scenes and motivating studies on effect on robotic tasks when such scene properties are varied. Further, our dataset and tools allow a correspondence between the real scenes used for creating the dataset and the rendered synthetic scenes, which motivates their use to create testbeds for studies in sim-to-real transfer. Similar to Figure 15 of the main paper, we show several examples of such capabilities enabled by OpenRooms in Figure \ref{fig:robotics_qualitative_examples}.

\vspace{-0.4cm}
\paragraph{Navigation}
We provide a simple example to show the support for navigation tasks. In this example, a two-wheeled Turtlebot is asked to navigate in an indoor room, from a starting location to a target location. The agent has a three-dimensional state space $S$ and a four-dimensional continuous action space $A$. The state $s \in S$ is the agent's 3D position. The first two dimensions of the action space $A$ correspond to moving forward or backward for an non-negative distance $d$. The other two dimensions represent turning left or right for an angle within $[-\pi, \pi]$. The robot is given a sequence of actions $\{a_1, a_2, \dots, a_T\}$ to accomplish the navigation task. Figure \ref{fig:navigation} shows a few frames from the resulting video. Besides directly working on the robot's 3D positions, we also provide a variety of observation modes, RGB images, surface normals, depth images, semantic segmentations and joint level state space for studying the navigation problems from different perspectives. In particular, we note that OpenRooms may allow navigation studies under different material properties and lighting conditions.

\vspace{-0.4cm}
\paragraph{Pushing with different frictions}
We conduct pushing experiments to show how the friction coefficients associated with different materials impact the object's behavior given the same pushing force applied by a robot. With the experiment setup (Table \ref{table:push_setup}), a cabinet is placed at position $P_0^C$ and a Turtlebot is initialized at position $P_0^R$ in the world frame. To generate a horizontal pushing force to the cabinet, the robot moves towards it at a constant speed $v^R$ for time $t^F$. Then the robot keeps still for time $t^O$ and during this period of time, the cabinet will eventually stop due to the friction between the cabinet and the floor. 

We perform this pushing task with two different floor materials while keeping the other conditions the same. Table \ref{table:push_physics} summarizes the physical properties of the objects appearing in the two scenarios. Table \ref{table:push_distance} compares the cabinet's position offset with respect to its starting position after being pushed by the robot on the carpet and on the wax floor, respectively. Figure \ref{fig:pushing} shows initial and final snapshots from the simulation videos of the pushing tasks.

Given the floor's material information, we compute the per-pixel friction coefficient map from the albedo and the roughness images, according to the method in Section 4.3 of the main paper and Section \ref{sec:friction} of the supplementary. The average of the resulting map is then fed into the physics engine as the floor's friction coefficient. Note that although not yet supported by popular physics simulators such as PyBullet, our dataset provides ground truth for spatially-varying friction coefficients, which can be incorporated into higher quality simulators in the future.

\begin{table}[t]
\small
\centering
\begin{tabular}{|l|l|} 
\thickhline
\multicolumn{2}{|c|}{Experimental setup} \\ 
\thickhline
 Gravity ($m/s^2$) & (0,0,-9.8) \\
 Robot mass (kg) &  0.27\\ 
 Cabinet mass (kg) & 2.00 \\
 Robot initial position $P_0^R$ (m) & (0.5,0,0) \\
 Cabinet initial position $P_0^C$ (m) & (0,0,0) \\
 Force exerting time $t^F$ (s)  & 0.08 \\
 Observing time $t^O$ (s)  & 1.67 \\
 Robot moving speed $v^R$ (m/s) & (2.4,0,0)  \\
 Robot moving distance (m) & 0.20 \\
\hline
\end{tabular}
\caption{Experimental setup for the pushing tasks.}
\label{table:push_setup}
\end{table}

\begin{table}[t]
\small
\centering
\begin{tabular}{|c|c|c|} 
 \thickhline
 object & material & friction coefficient \\
 \thickhline
 cabinet & wood & 0.76 \\ 
 floor 1 & carpet & 0.76 \\ 
 floor 2 & wax & 0.31 \\
 \hline
\end{tabular}
\caption{Physical properties of the objects involved in the pushing tasks. The friction coefficients are in the range [0,1].}
\label{table:push_physics}
\end{table}

\begin{table}[t]
\small
\centering
\begin{tabular}{|c|c|} 
 \thickhline
 scenario & object position offset (meter)  \\
 \thickhline
 Pushing on floor 1 & 0.12 \\ 
 Pushing on floor 2 & 0.23 \\
 \hline
\end{tabular}
\caption{Comparison of results on the pushing tasks with different friction coefficients for the floor.}
\label{table:push_distance}
\end{table}

\section{Segmentation, Multi-Tasking, Adaptation}
\label{sec:segmentation}

In this section, we provide: {\bf (a)} further qualitative results for semantic segmentation, {\bf (b)} results for instance segmentation, {\bf (c)} quantitative and qualitative results for multi-task shape, material and semantics estimation, {\bf (d)} domain adaptation for depth estimation.

\begin{figure*}
\centering
\includegraphics[width=0.9\textwidth]{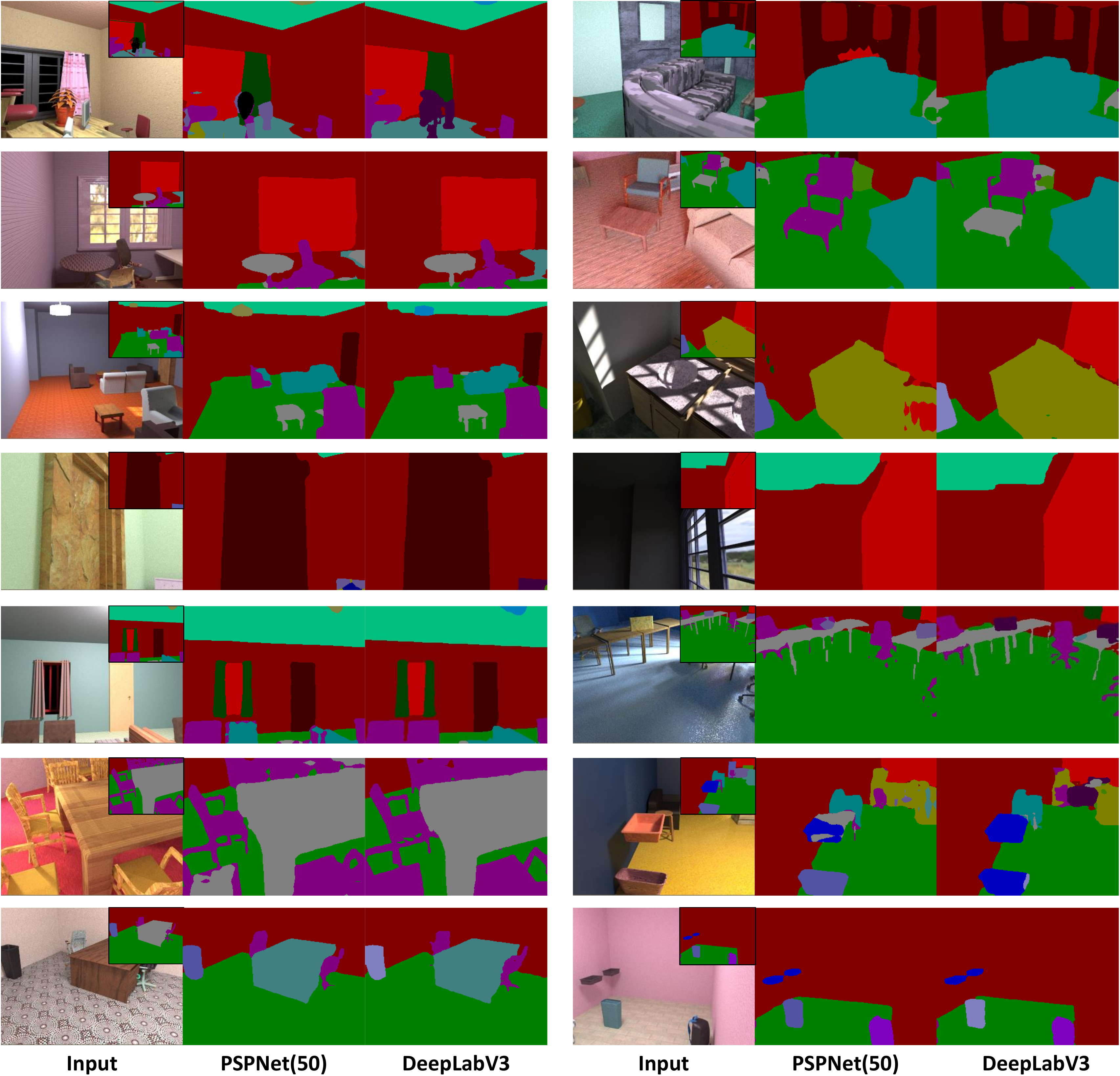}
\vspace{-3mm}
\caption{Further qualitative examples of semantic segmentation on OpenRooms.}
\label{fig:semseg-more}
\vspace{-0.4cm}
\end{figure*}

\begin{figure*}
\centering
\includegraphics[width=0.9\textwidth]{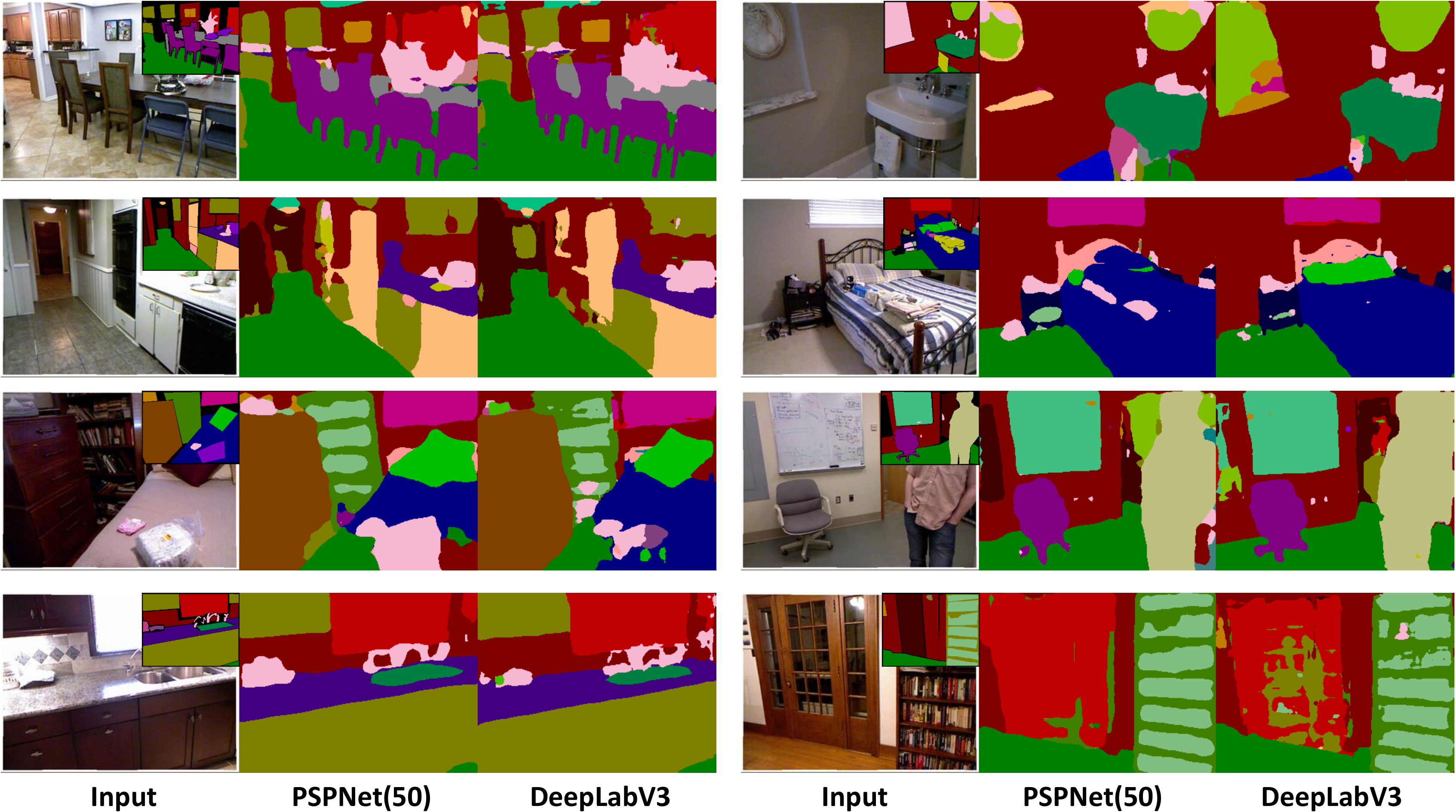}
\vspace{-3mm}
\caption{Further qualitative examples of semantic segmentation on NYUv2.}
\label{fig:semseg-more-nyu}
\vspace{-0.4cm}
\end{figure*}

\vspace{-0.4cm}
\paragraph{Semantic segmentation}
The main paper provides quantitative numbers and some visualizations for semantic segmentation networks trained on OpenRooms data. More qualitative results for semantic segmentation using PSPNet(50) and DeepLabV3 evaluated on OpenRooms test set and the NYUv2 test set are in Figures \ref{fig:semseg-more} and \ref{fig:semseg-more-nyu}, respectively. 

\begin{figure}
\centering
\includegraphics[width=\linewidth]{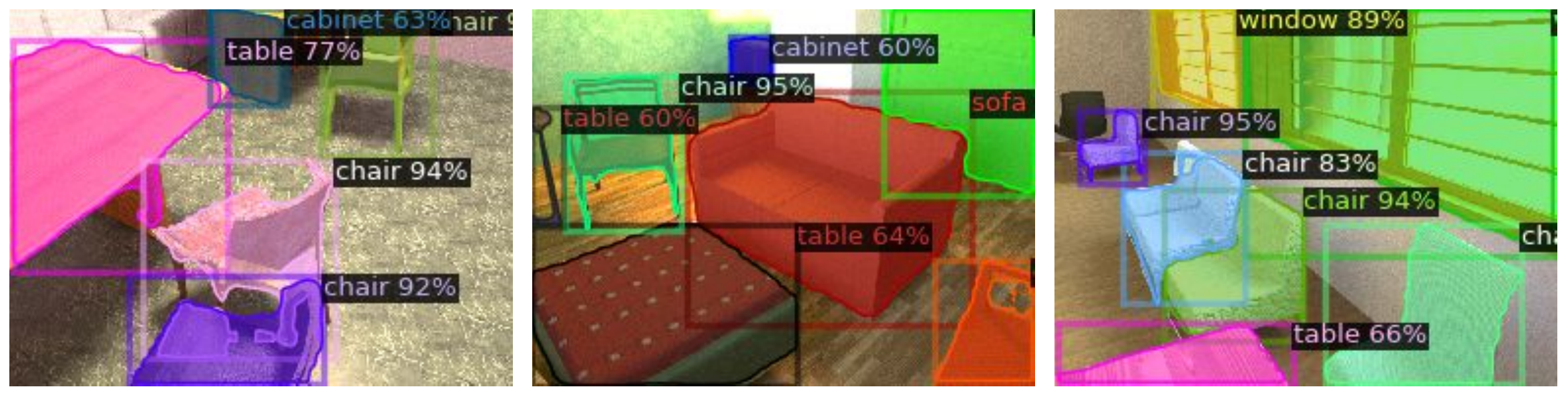}
\vspace{-6mm}
\caption{Qualitative examples of instances segmentation results on Openrooms.}
\label{fig:instance_sup}
\end{figure}

\vspace{-0.4cm}
\paragraph{Instance segmentation}
The results of instance segmentation on OpenRooms, using the same network architecture as light source detection, are shown in Table \ref{tab:instance40} for the categories that overlap with the NYU label space. A few qualitative examples are shown in Figure~\ref{fig:instance_sup}, indicating that the proposed dataset may also be useful for studies in instance segmentation. 


\begin{table}
\scriptsize
\addtolength{\tabcolsep}{-1.1pt}  
\begin{center}
\begin{tabular}{ |c|cc| }
  \hline
    &  bbox &  seg  \\

   \hline
    AP(0.5:0.95)&  39.1 &  48.464 \\
    \hline
    AP-cabinet&  34.00  &  52.97  \\
  AP-bed&  56.63 &  66.33  \\
  AP-chair&  43.89 &  48.10  \\
AP-sofa&  51.19 &  61.29  \\
AP-table&  43.86 &  53.27  \\
AP-door&  63.52 &  75.74  \\
AP-window&  42.81 &  65.53  \\
AP-bookshelf&  46.07 &  53.24  \\
AP-counter&  6.92 &  7.50  \\
AP-desk&  13.39 & 23.39 \\
AP-curtain&  41.27 &  35.05  \\

AP-bathtub&  58.55 &  62.98  \\
AP-bag&  16.52 &  52.02  \\
AP-otherstructure&  1.01 &  1.19 \\
AP-otherfurniture&  60.52 &  67.78   \\
AP-otherprop&  51.81 &  59.83 \\

 \hline
\end{tabular}
\vspace{-0.2cm}
\caption{Instance segmentation results on the OpenRooms dataset using the label space of NYU.}
\label{tab:instance40}
\end{center}
\addtolength{\tabcolsep}{1.1pt}  
\vspace{-0.5cm}
\end{table}

\begin{figure*}
\centering
\includegraphics[width=0.9\textwidth]{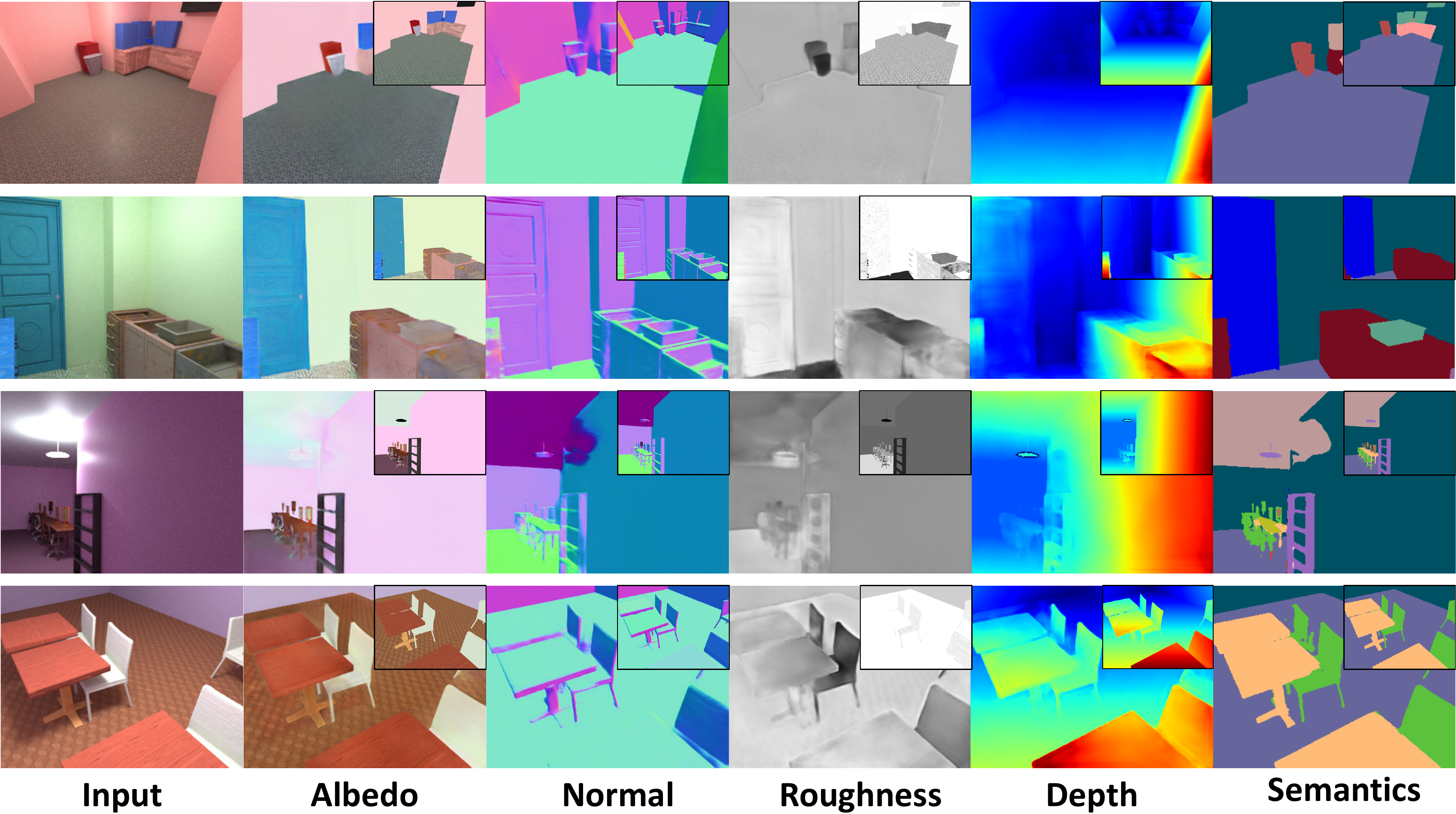}
\vspace{-3mm}
\caption{Further qualitative examples of multi-task estimation on OpenRooms.}
\label{fig:multitask_sup}
\end{figure*}

\begin{figure*}
\centering
\includegraphics[width=0.9\textwidth]{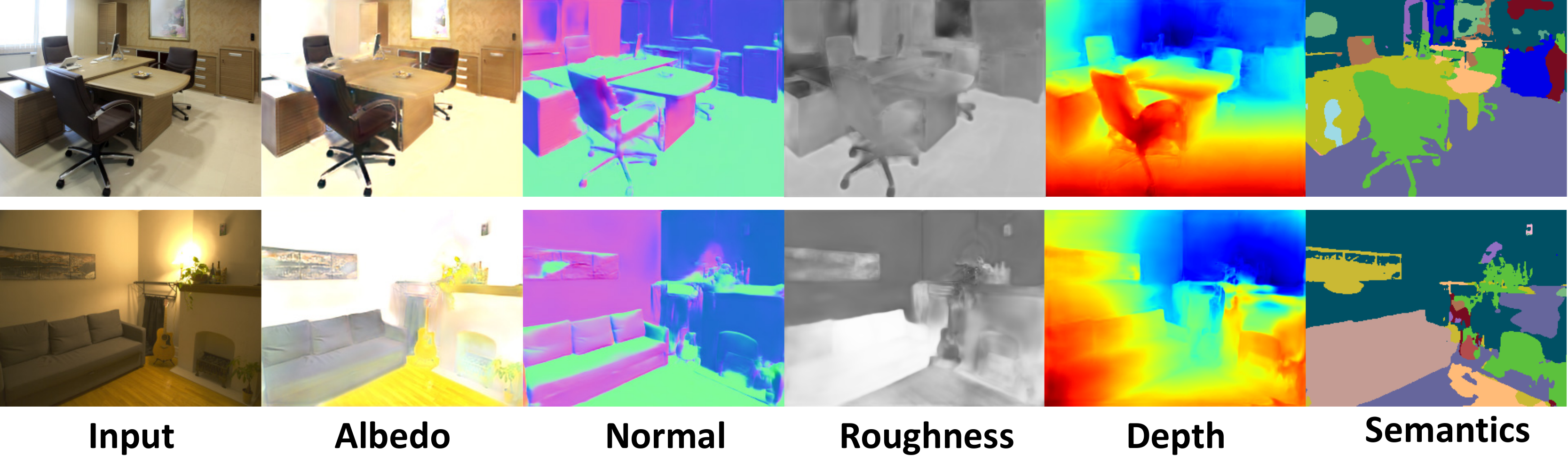}
\vspace{-3mm}
\caption{Further qualitative examples of multi-task estimation on real images.}
\label{fig:multitask-real}
\end{figure*}

\vspace{-0.4cm}
\paragraph{Multi-task learning}
We quantitatively evaluate our multi-task model (Cascade0) with estimation of albedo, normal, depth, roughness as well as semantic segmentation on the test set of OpenRooms, reporting the results in Table \ref{tab:multi_task}. As compared to the original model which does not include a segmentation branch, our multi-task model provide competitive results to albedo, normal, depth and roughness estimation while enable additional semantics estimation with reasonable performance. Besides the qualitative results in Figure 12 of the main paper, we provide further qualitative results of the multi-task model in Figure~\ref{fig:multitask_sup} for OpenRooms test images and Figure~\ref{fig:multitask-real} for real images.
 
\begin{table*}[t]
\centering

\begin{tabular}{|c|c|c|c|c|c|c|c|c||c|c|}
\hline
\multirow{2}{*}{} & Albedo & \multicolumn{3}{c|}{Normal} & \multicolumn{3}{c|}{Depth} & Roughness & \multicolumn{2}{c|}{Semantics}\\
\cline{2-11}
 & loss & loss & mean($^{\circ}$) & med.($^{\circ}$) & loss & Abs Rel & RMSE & loss & mIoU & mAcc\\
\hline
Multi-task model & 9.47  & 4.08 & 14.17 & 4.90 & 2.98 & 0.1066 & 0.2647 & 6.69 & 23.1 & 28.8\\
\hline
Segmentation-only & - & - & - & - & - & - & - & - & 23.4 & 28.9 \\
\hline 
W/o segmentation & 8.66  & 4.12 & 14.32 & 5.19 & 3.15 & 0.1070 & 0.2573 & 6.33 & - & -\\
\hline
\end{tabular}
\caption{Ablation study for models including the multi-task model, the segmentation-only model where the albedo, normal, depth and roughness heads are removed, as well as the model without the segmentation head while keeping all other four heads, evaluated on the OpenRooms dataset. We report the scale invariant L2 loss ($10^{-3}$) for Albedo, L2 loss ($10^{-2}$) for normal, scale invariant $\log$ L2 loss ($10^{-2}$) for depth, L2 loss ($10^{-2}$) for roughness and scale invariant $\log(x+1)$ L2 loss for per-pixel lighting. Angular errors for normal estimation, Abs Rel and RMSE errors for depth estimation (valid depth range of 0.1m -- 8m), as well as mean IoU (mIoU) and mean accuracy (mAcc) for semantic segmentation are also included.}
\label{tab:multi_task}
\end{table*}

\begin{figure}
\centering
\includegraphics[width=\columnwidth]{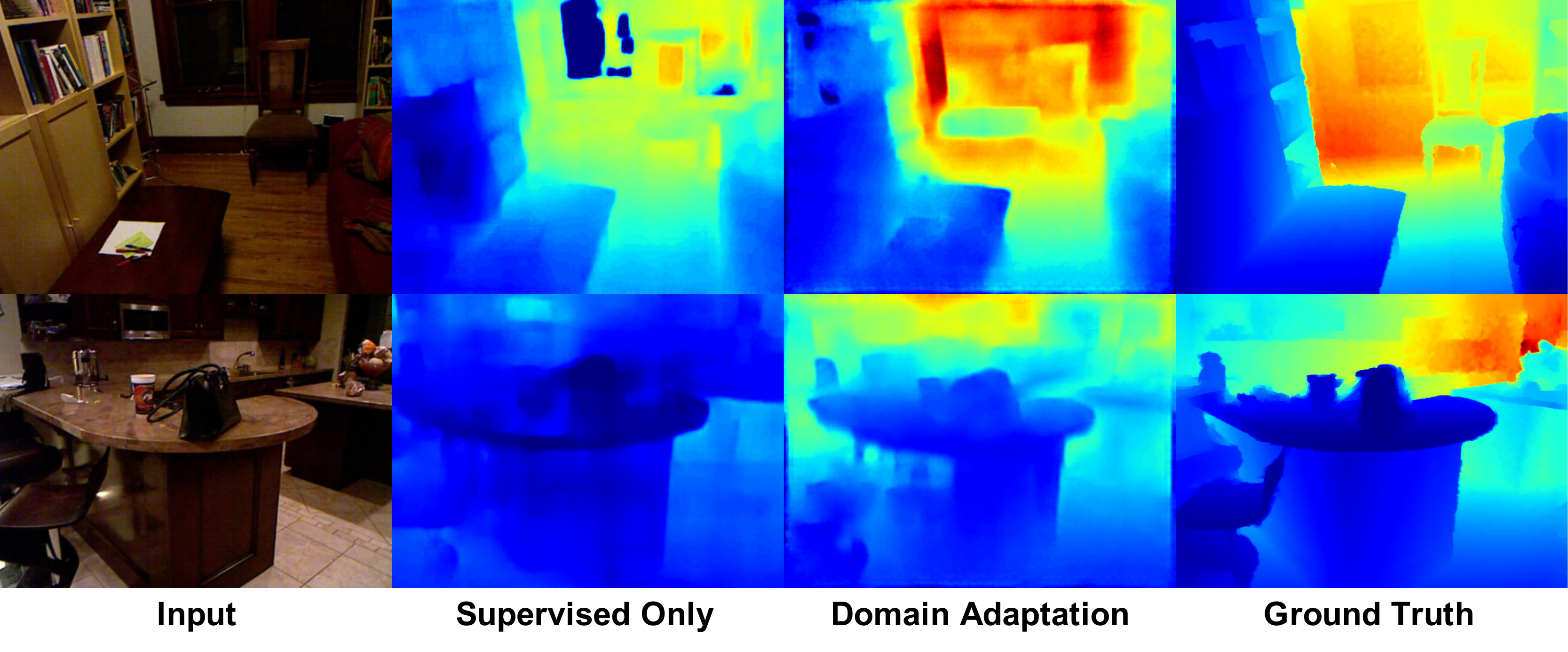}
\vspace{-6mm}
\caption{Qualitative results of depth estimation on Supervised only model and Domain Adaptation model.}
\label{fig:da_depth}
\vspace{-0.4cm}
\end{figure}

\vspace{-0.4cm}
\paragraph{Domain Adaptation for depth estimation}
The availability of large-scale synthetic data with labels also allow studies where labels from OpenRooms may be used for domain adaptation to real scenes where labels might not be available. In this section, we show an example for depth estimation, where OpenRooms is the source domain with 100k images, while the target domain is unlabeled NYU with 15k frames from its raw dataset. 
We train an unsupervised domain adaptation model for depth estimation using T2Net~\cite{Zheng_2018_ECCV}. We clip the depth values within the range of $[0, 10]$ meters and evaluate the predictions within the range of $[1, 8]$ meters, following the settings of \cite{Zheng_2018_ECCV}. We observe that the domain gap can be largely addressed with unsupervised domain adaptation in Table~\ref{tab:da_depth} and Figure~\ref{fig:da_depth}. The target supervised numbers are trained on the NYUv2 training set which is composed of $1440$ frames. Further, pre-training on the large-scale OpenRooms synthetic data before fine-tuning on the smaller scale NYUv2 training data leads to improved performance.

\begin{table*}[t]
\centering
\begin{tabular}{l|cccc|ccc|}
\cline{2-8}
& \multicolumn{4}{c|}{lower is better} & \multicolumn{3}{c|}{higher is better}\\ 
\cline{2-8} 
& Abs Rel & Sq Rel & RMSE   & RMSE log & $\delta$ < $1.25$ & $\delta$ < $1.25^2$ & $\delta$ < $1.25^3$ \\ 
\hline
\multicolumn{1}{|l|}{Source Only}  & 0.4723  & 1.0160 & 1.8520 & 0.7765   & 0.2049  & 0.3806 & 0.5379 \\ 
\hline
\multicolumn{1}{|l|}{Unsupervised Domain Adaptation} & 0.2388 &    0.3060 &    0.9430 &    0.3280 &    0.5770 &    0.8266 &    0.9336 \\ 
\hline
\multicolumn{1}{|l|}{Target Only} & 0.2031 &    0.2139 &    0.7477 &    0.2603 &    0.6536 &    0.8968 &    0.9675 \\ 
\hline
\multicolumn{1}{|l|}{Source + Target Finetuned} & 0.1721 &    0.1865 &    0.6663 &    0.2206 &    0.7465 &    0.9261 &    0.9753 \\ 
\hline
\end{tabular}
\caption{Unsupervised domain adaptation results for depth estimation on NYUv2 labeled test set. \textit{Source Only} model is trained on OpenRooms while \textit{Unsupervised Domain Adaptation} model is trained on OpenRooms and adapted to NYU unlabeled data. \textit{Target Only} model is trained on labeled NYUv2. \textit{Source + Target Finetuned} model uses the Source Only model as pre-trained weight and finetuned on labeled NYUv2. Metrics are computed using the labeled data which does not appear during training.}
\label{tab:da_depth}
\end{table*}

\section{Dataset Creation using SUNRGBD Data}
\label{sec:SUNRGBD}

\begin{figure*}[ht]
\centering
\includegraphics[width=0.95\textwidth]{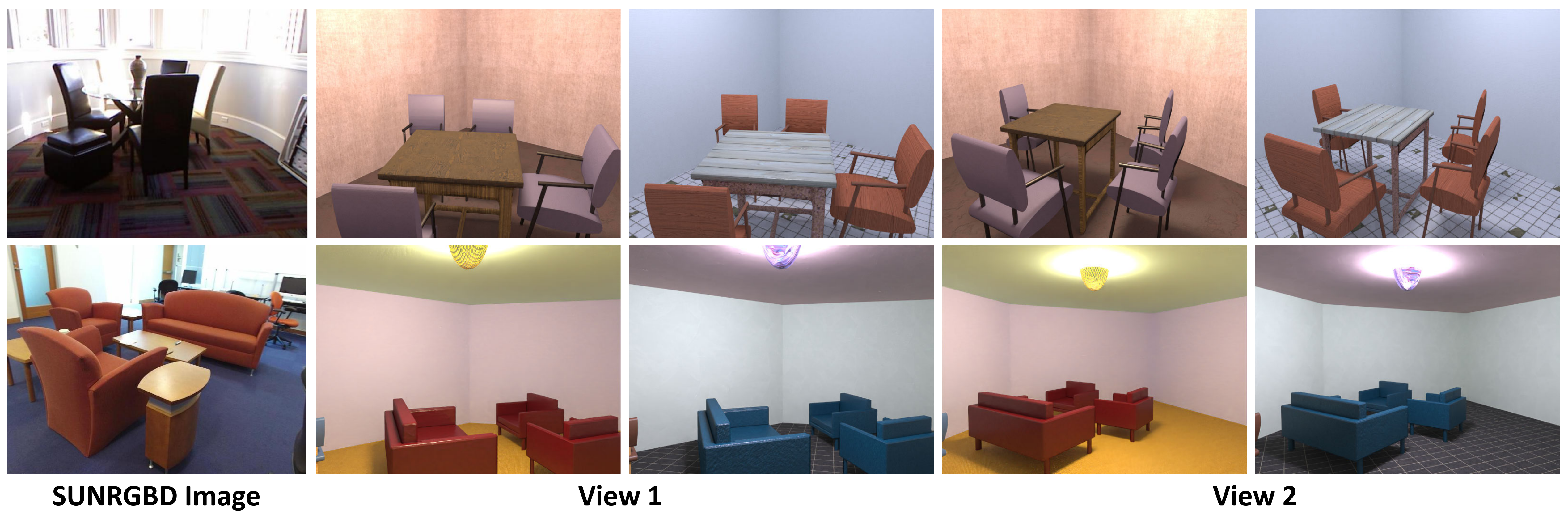}
\caption{Synthetic scene reconstruction results using scanned indoor scenes from SUNRGBD dataset. We visualize the reconstructed scenes rendered from different views with different material assignments.  }
\vspace{-0.4cm}
\label{fig:SUNRGBD}
\end{figure*}

To demonstrate that our framework can generalize to other datasets, we present our scene reconstruction results based on scanned indoor scenes from the SUNRGBD dataset. Unlike ScanNet \cite{dai2017scannet}, SUNRGBD only contains partial scans of the rooms with extremely incomplete and sparse point clouds. Moreover, unlike Scan2CAD \cite{avetisyan2019scan2cad}, SUNRGBD only has 3D bounding box annotations for furniture locations and lacks full poses. Using this as initialization, we adjust the pose of the CAD models by simply using grid search to minimize the Chamfer distance between the CAD model and the point cloud in the bounding box. Then we assign appropriate materials and lighting to the CAD models, as described in the main paper. In our experience, differing qualities of scans need to be addressed for geometry creation in different datasets, but our material and lighting mapping transfer across datasets with minimal effort. In Figure \ref{fig:SUNRGBD}, we visualize the reconstruction results for SUNRGBD by rendering the created scenes from different viewpoints, with different material assignments. The rendered images present diverse appearances with plausible material and lighting assignments, with complex visual effects such as soft shadows and specularity being correctly handled. 

\section{Microfacet BRDF Model}
\label{sec:BRDF}

We use the simplified microfacet BRDF model of \cite{karis2013unreal}. Let $A$, $N$, $R$ be the diffuse albedo, normal and roughness. Our BRDF model $\mathbf{f}(A, N, R)$ is defined as 
\begin{align*}
\!\!\!\! \mathbf{f}(A, N, R, l, v) &= \frac{A}{\pi} + 
\frac{\mathbf{D}(h, R)\mathbf{F}(v, h)\mathbf{G}(l, v, N, R)}{4(N \cdot l)(N\cdot v)} \\
\!\!\!\! \mathbf{D}(h, R) &=  \frac{R^{4} }{\pi((N\cdot h)^{2}(R^{4}-1) + 1)^{2} } \\
\!\!\!\! \mathbf{F}(v, h) &=  (1 - F_{0})2^{(-5.55473(v \cdot h) - 6.98316)v \cdot h}\! + F_0 \\
\!\!\!\! \mathbf{G}(l, v, N, R) &=  \mathbf{G}_{1}(v, N)\mathbf{G}_{1}(l, N) \\
\!\!\!\! \mathbf{G}_{1}(v, N)  &=  \frac{N \cdot v}{(N \cdot v)(1-k) + k},~~ k = \frac{(R + 1)^{2}}{8}
\end{align*}
where $v$ and $l$ are the view and light directions, while $h$ is the half angle vector. Further, $\mathbf{D}(h, R)$, $\mathbf{F}(v, h)$ and $\mathbf{G}(l, v, N, R)$ are the distribution, Fresnel and geometric terms, respectively. We set $F_{0}=0.05$, following \cite{karis2013unreal}.

\section{Ground Truth for Lighting}
\label{sec:lighting}

Ground truth for complex light transport effects is difficult to acquire in real scenes and not available in prior synthetic datasets. Compared to prior datasets for indoor lighting estimation such as \cite{li2020inverse}, OpenRooms not only provides spatially-varying per-pixel environment maps but also extensive ground truth for light source positions, colors and intensities.
Moreover, it also provides ground truth for individual contributions of each light source to the pixel intensity, rendered with direct illumination and with direct and indirect illumination combined, with and without occlusion being considered. Such new supervisions allow us to model the scene appearance with light sources in the scene turned off or on, which may enable new challenging lighting editing applications in the future. Figure 5 in the main paper and Figure \ref{fig:lightSource_supp} in the supplementary show examples of our light source supervision. We now provide further implementation details.  

\vspace{-0.4cm}
\paragraph{Per-light shading} To render the individual contribution of each light source in the scene, we need to turn all other light sources off and keep only one light source on. This is straightforward for lamps, but not for windows, especially if there are multiple windows in the room, as shown in Figure \ref{fig:lightSource_supp}. To achieve this, we provide the plane parameters of each window to the renderer. When sampling the environment map, we check whether the ray hitting the environment map passes through the plane approximation of the window geometry. We only consider the contributions of those rays that pass through the window. 

\vspace{-0.4cm}
\paragraph{Shading without occlusion} 
We render all the ground-truth with our customized OptiX-based GPU renderer. 
OptiX handles visibility term by calling its $\mathtt{rtTrace}$ function to detect if a ray will be occluded. To render without the visibility term, we simply do not use $\mathtt{rtTrace}$ in the renderer. 

\begin{figure*}[tbh]
\centering
\includegraphics[width=\textwidth]{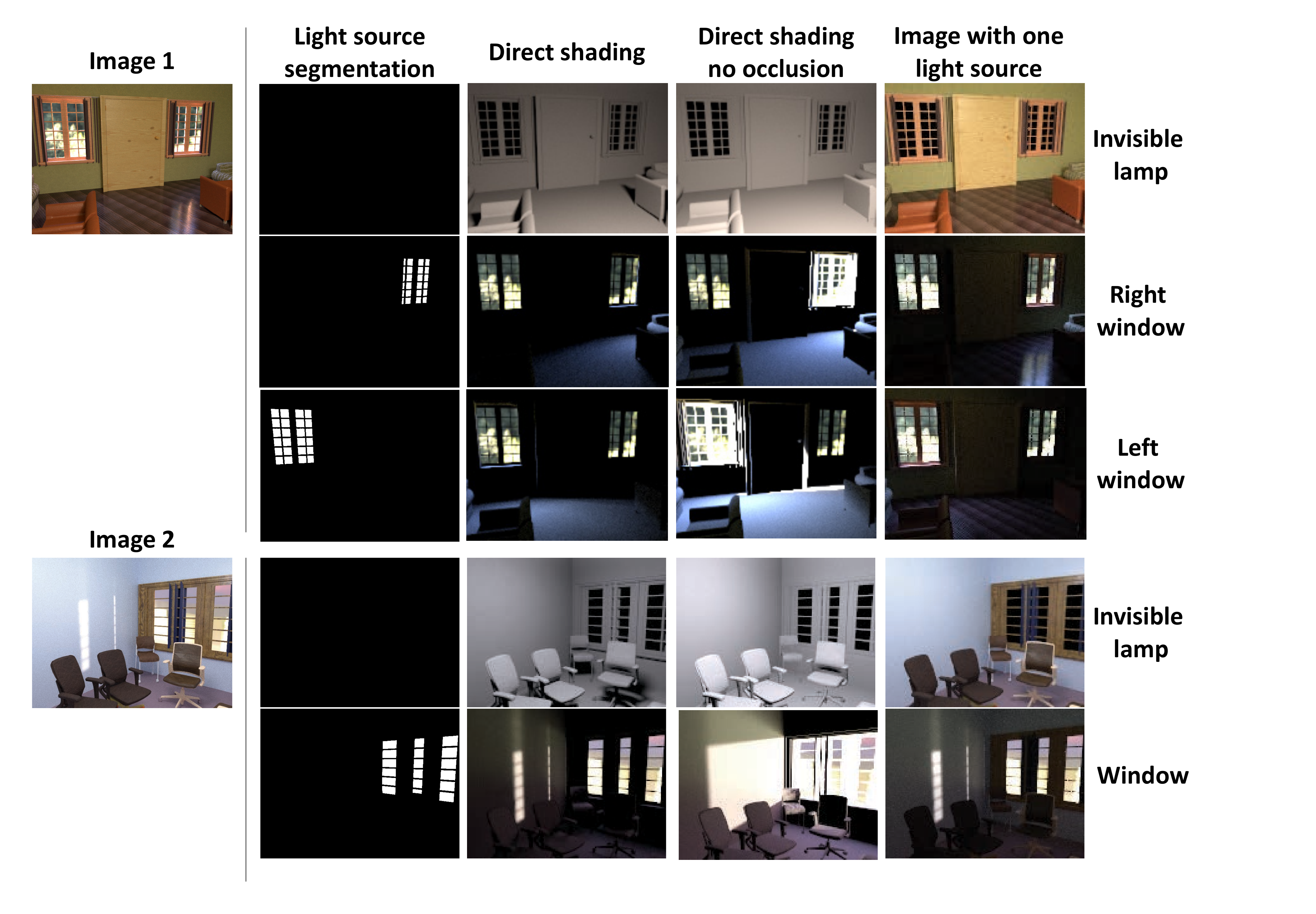}
\caption{Light source ground-truth that can be provided by OpenRooms dataset. Unlike prior dataset \cite{li2020inverse}, we provide the contribution of each individual light source, with the influences of direct/indirect illumination and visibility being separated. }
\label{fig:lightSource_supp}
\end{figure*}

\section{Physically-Based GPU Renderer}
\label{sec:renderer}

We render our dataset efficiently using a physically-based GPU renderer. We make one crucial design choice to improve the rendering speed while maintaining the rendering quality -- when rendering the spatially-varying lighting, we not only uniformly sample the hemisphere, but also sample the light sources. The contributions of the two sampling methods can be combined together using the standard power rule in multiple importance sampling \cite{veach1997robust}. This allows us to capture the radiance from small light sources in the scene with far fewer samples. More formally, let $\eta$ be the ray direction, $\mathbf{P}_{L}(\eta)$ be the probability of sampling $\eta$ when sampling the light sources, $\mathbf{P}_{U}(\eta)$ be the probability of uniformly sampling the hemisphere and $\mathcal{I}$ be an indicator function that is equal to $1$ when a light source is sampled and $0$ otherwise. Further, let $\mathbf{L}(\eta)$ be the radiance. Then, the contribution of sampling $\eta$ towards the corresponding pixel on the hemisphere can be written as:
\begin{equation}
I \cdot \frac{\mathbf{P}^{2}_{L}}{\mathbf{P}^{2}_{L} + \mathbf{P}^{2}_{U}} \frac{\mathbf{L} }{\mathbf{P}_{L} } + (1 -I) \cdot \frac{\mathbf{P}^{2}_{U}}{\mathbf{P}^{2}_{L} + \mathbf{P}^{2}_{U}} \frac{\mathbf{L} }{\mathbf{P}_{U}},
\end{equation}
where dependence of $\mathbf{L}$, $\mathbf{P}_{L}$, $\mathbf{P}_{U}$ on $\eta$ is omitted for clarity.

{\small
\bibliographystyle{ieee_fullname}
\bibliography{egbib}
}

\end{document}